\PassOptionsToPackage{sort&compress}{natbib}
\PassOptionsToPackage{table}{xcolor}

\documentclass{article}

\usepackage[preprint]{neurips_2025}

\usepackage[utf8]{inputenc} 
\usepackage[T1]{fontenc}    
\usepackage{hyperref}       
\usepackage{url}            
\usepackage{booktabs}       
\usepackage{amsfonts}       
\usepackage{nicefrac}       
\usepackage{microtype}      
\usepackage{pifont}
\usepackage{MnSymbol}
\usepackage{accents}
\usepackage{cancel}
\usepackage{graphicx}
\usepackage{wrapfig}
\usepackage{caption}
\usepackage{subcaption}
\usepackage{tabularx}
\usepackage{array}
\usepackage{float} 
\usepackage[most]{tcolorbox}
\usepackage{multirow}
\usepackage{algorithm}
\usepackage{algorithmic}
\usepackage{xspace}
\usepackage{enumitem}
\usepackage{cleveref}

\definecolor{darkblue}{rgb}{0, 0, 0.5}
\hypersetup{colorlinks=true, citecolor=darkblue, linkcolor=darkblue, urlcolor=darkblue}
\newcommand*\titleimage[1]{\raisebox{-0.35\baselineskip}{\includegraphics[height=1.2\baselineskip]{#1}$\,$}}

\newcommand{\methodtitleemoji}{\titleimage{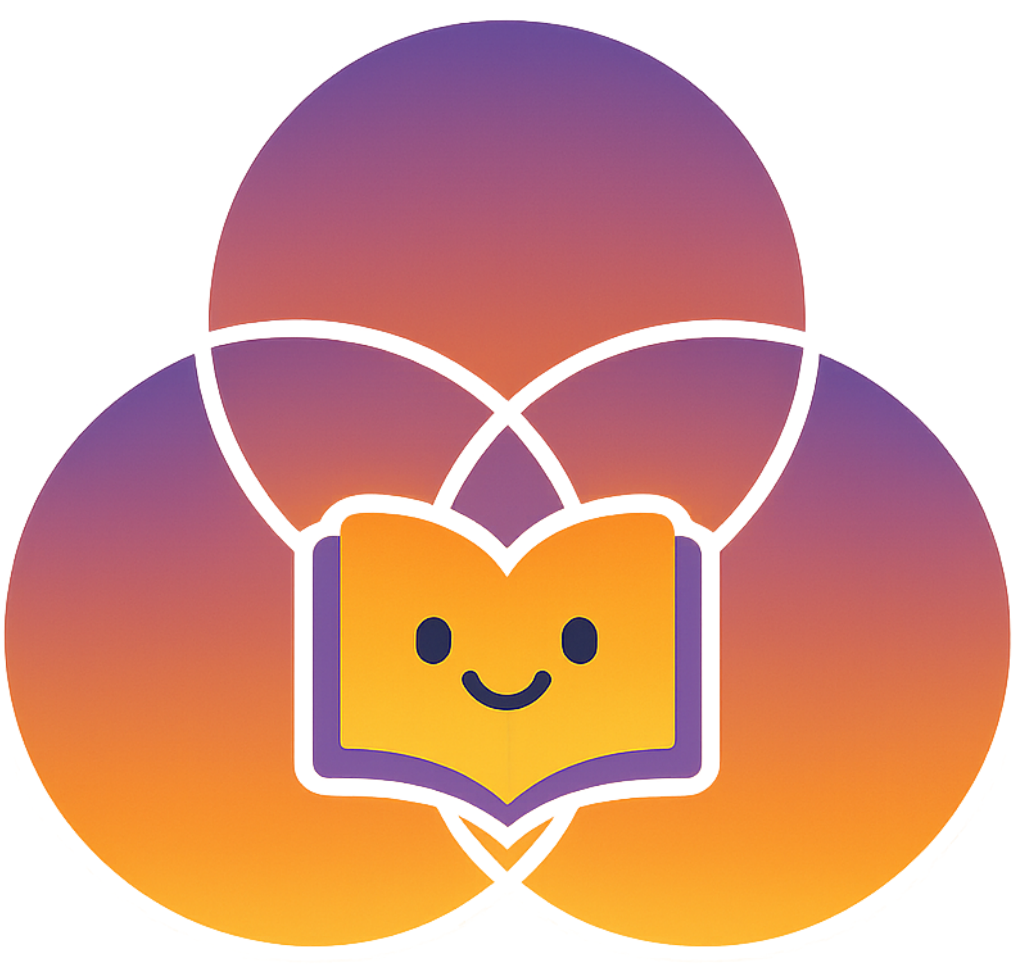}}
\DeclareRobustCommand{\methodemoji}{%
    \texorpdfstring{\protect\raisebox{-0.3\baselineskip}{\protect\includegraphics[height=1.2\baselineskip]{figures/MSMU.pdf}}\hspace{0.2em}}{}%
}

\title{\methodtitleemoji DUSK: Do Not Unlearn Shared Knowledge}

\newcommand{\hlorange}[1]{\begingroup
  \setlength{\fboxsep}{1.2pt}%
  \colorbox{orange!70}{\textcolor{black}{#1}}%
\endgroup}

\newcommand{\hlblue}[1]{\begingroup
  \setlength{\fboxsep}{1.2pt}%
  \colorbox{blue!50}{\textcolor{black}{#1}}%
\endgroup}

\author{
Wonje Jeung$^{1}$\thanks{Equal contribution.}\hspace{0.5em}
Sangyeon Yoon$^{2}$$^*$\hspace{0.5em}
Hyesoo Hong$^{1}$\hspace{0.5em}
Soeun Kim$^{1}$\hspace{0.5em}\\
\textbf{Seungju Han}$^{3}$\hspace{0.5em}
\textbf{Youngjae Yu}$^{1}$\hspace{0.5em}
\textbf{Albert No}$^{1}$\thanks{Corresponding author}\vspace{0.5em}\\
{\hspace{0.5em}$^1$Yonsei University\hspace{0.5em}$^2$Hongik University\hspace{0.5em}$^3$Standford University} \vspace{0.5em}}

\begin{document}

\maketitle

\begin{abstract}
Large language models (LLMs) are increasingly deployed in real-world applications, raising concerns about the unauthorized use of copyrighted or sensitive data. Machine unlearning aims to remove such “forget” data while preserving utility and information from the “retain” set. However, existing evaluations typically assume that forget and retain sets are fully disjoint, overlooking realistic scenarios where they share overlapping content. For instance, a news article may need to be unlearned, even though the same event—such as an earthquake in Japan—is also described factually on Wikipedia. Effective unlearning should remove the specific phrasing of the news article while preserving publicly supported facts.
In this paper, we introduce \methodemoji DUSK, a benchmark designed to evaluate unlearning methods under realistic data overlap. DUSK constructs document sets that describe the same factual content in different styles, with some shared information appearing across all sets and other content remaining unique to each. When one set is designated for unlearning, an ideal method should remove its unique content while preserving shared facts. We define seven evaluation metrics to assess whether unlearning methods can achieve this selective removal.
Our evaluation of nine recent unlearning methods reveals a key limitation: while most can remove surface-level text, they often fail to erase deeper, context-specific knowledge without damaging shared content.
We release DUSK and code at \url{https://ai-isl.github.io/dusk} to support precise, reliable unlearning for real-world use.
\end{abstract}

\section{Introduction}

Large language models (LLMs) are typically trained on web-scale corpora that include copyrighted materials, personal data, and user-generated content~\citep{carlini2021extracting,nasr2023scalable}. As these models are deployed in real-world applications, individuals and organizations increasingly demand the removal of specific training examples due to legal and ethical concerns. These demands are driven by privacy regulations such as the GDPR~\citep{voigt2017eu} and reinforced by recent lawsuits~\citep{grynbaum2023times,openailawsuit2,githublitigation} over the unauthorized use of proprietary content. This has led to growing interest in machine unlearning~\citep{nguyen2022survey,liu2025rethinking}, which focuses on removing the influence of \textit{forget data} (e.g., copyrighted documents) from a trained model without retraining from scratch, while preserving information from \textit{retain data}.
To evaluate unlearning algorithms, several benchmarks have recently been proposed~\citep{maini2024tofu, shi2024muse, jin2024rwku}. For example, MUSE~\citep{shi2024muse} targets copyright-related scenarios by focusing on the removal of entire documents. In such cases, unlearning algorithms are expected to erase both verbatim text and the underlying knowledge from the forget set, while preserving information learned from the retain set.

However, existing evaluations often assume that the forget set and retain set are disjoint, overlooking the complexity of real-world data. In practice, documents frequently contain overlapping information. For instance, a New York Times report might state, "A 6.2 magnitude earthquake struck Tokyo on Monday," while Wikipedia might describe the same event as "A strong tremor shook the Japanese capital at the start of the week." Despite the different phrasing, both convey the same core facts—location, magnitude, and timing. This highlights a critical challenge: effectively unlearning only the unique information from a forget set without disrupting widely supported facts in the retain set.

\begin{figure}[t!]
\centering
\includegraphics[width=0.93\textwidth]{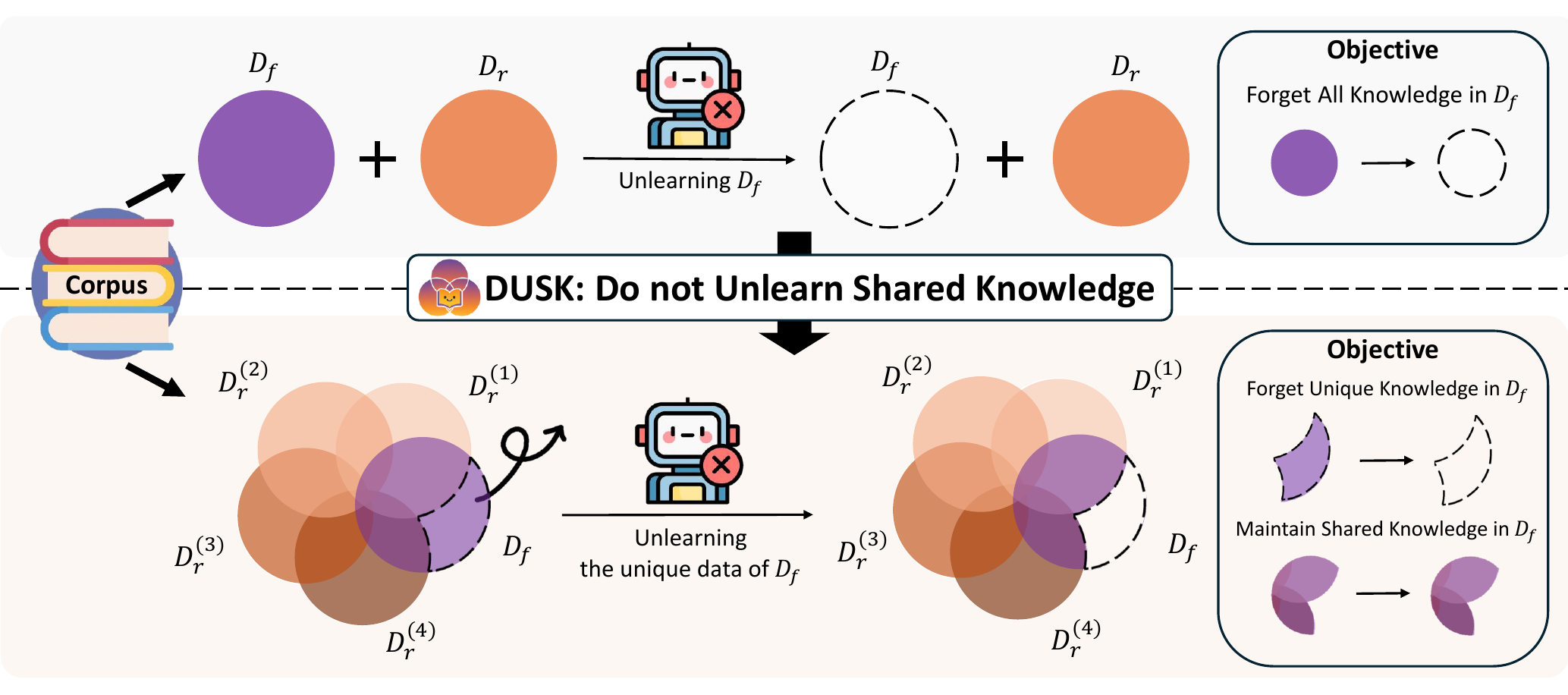}
\vskip -0.5em
\caption{\methodemoji \textbf{DUSK provides a realistic unlearning evaluation scenario where forget documents ($\mathcal{D}_f$) contain both unique information to be forgotten and shared knowledge that must be preserved.} Unlike conventional setups that naively erase entire forget sets, DUSK evaluates whether unlearning methods can selectively remove sensitive information while retaining shared knowledge supported by other documents in the retain set ($\mathcal{D}_r$), which is not subject to forgetting.} \label{fig:overview}
\end{figure}

To address this gap, we introduce \methodemoji DUSK, a benchmark for evaluating unlearning in realistic multi-source settings where the same information can appear across both forget and retain data. DUSK constructs document sets that describe identical factual content in different styles, enabling controlled attribution: some information is present only in the forget set, while other content remains supported by the retain set. Specifically, DUSK is constructed using 120 synthetic professor profiles, organized into five documents. Each document contains 12 unique profiles that appear only in that document, along with 60 shared profiles that appear across multiple documents. This structure allows systematic evaluation of whether an algorithm can effectively remove content uniquely attributable to the forget set without erasing widely supported, shared knowledge. An overview of the DUSK framework is presented in~\Cref{fig:overview}.
DUSK defines seven evaluation metrics: (1) Verbatim Memorization, which checks whether exact text from the forget set has been fully removed; (2) Unique Forget Knowledge, which measures whether forget-only content is effectively erased; (3) Shared Knowledge, which ensures that overlapping information is preserved; (4) Unique Retain Knowledge, which verifies that retain-only facts remain intact; (5) Downstream Capability, which evaluates whether the model maintains general utility after unlearning; (6) Privacy Leakage, which checks for residual leakage of forget set information; and (7) Retain Deviation, which assesses whether the model's behavior on the retain set is preserved.

We evaluate nine unlearning methods on the DUSK benchmark. Our results reveal a critical limitation of existing methods: while these methods aim to remove content uniquely attributable to the forget set, they often struggle to fully separate this content from information still supported by the retain set. This incomplete disentanglement can degrade shared knowledge, compromising the model's overall utility. These findings highlight a fundamental challenge in current unlearning approaches: precisely distinguishing forget-specific information from retained knowledge. To support further research, we release DUSK as a public benchmark for evaluating unlearning in real-world scenarios.

\section{Related Work}

\paragraph{Machine Unlearning in LLMs: Methods and Applications.}
Machine unlearning aims to selectively remove the influence of \textit{forget data} from a trained model while preserving its performance on \textit{retain data}~\citep{cao2015towards,brophy2021machine,guo2019certified,jeon2024information}.
Recent efforts have extended unlearning techniques to large language models (LLMs)~\citep{liu2025rethinking}, enabling their use in a range of applications such as removing copyrighted content~\citep{kassem2023preserving,wei2024evaluating}, eliminating sensitive or harmful knowledge~\citep{maini2024tofu,yousefpour2025representation,zhang2024safe}, mitigating bias~\citep{dige2024can,jeung2024large}, and performing model editing~\citep{guo2024mechanistic}.
Most methods achieve unlearning by fine-tuning on forget data~\citep{chen2023unlearn,jia2024soul,cao2015towards,barbulescu2024each}, commonly using gradient ascent~\citep{jang2022knowledge} or preference optimization~\citep{zhang2024negative}. To scale these methods to large models, recent work has explored approaches such as guardrail-based techniques~\citep{thaker2024guardrail,gaolarge}, and in-context unlearning~\citep{pawelczyk2023context}.
Despite this progress, recent studies have highlighted the fragility of current unlearning techniques~\citep{hu2024jogging,lynch2024eight,thaker2024position,zhang2024does,joshi2024towards,jeung2025seps}, revealing fundamental challenges in achieving robust and reliable unlearning in practice.

\paragraph{Machine Unlearning in LLMs: Benchmarks.}
As machine unlearning methods for LLMs evolve, the need for comprehensive evaluation benchmarks has become increasingly important. Early work introduced the ``Who is Harry Potter'' (WHP) task~\citep{eldan2023s}, which targets entity-specific forgetting by fine-tuning models on fictional corpora and evaluating unlearning through related prompts while monitoring retention on unrelated tasks.
Subsequent efforts expand the scope to hazardous knowledge, with the WMDP~\citep{li2024wmdp} focusing on removing information related to biosecurity and cybersecurity while preserving general model capabilities.
To enable controlled evaluation of unlearning, TOFU~\citep{maini2024tofu} constructs synthetic author profiles with associated question-answer pairs generated by GPT-4~\citep{achiam2023gpt}. This synthetic setup ensures that the model's knowledge of these authors originates solely from the fine-tuning process, allowing for precise assessment of unlearning effectiveness.
MUSE~\citep{shi2024muse} advances evaluation by systematically assessing unlearning algorithms across six dimensions, including forgetting effectiveness, privacy leakage, utility retention, scalability, and sustainability. It aims to remove both verbatim content and underlying knowledge within the forget set. CoTAEval~\citep{wei2024evaluating} instead focuses on a narrower goal—removing only verbatim memorization while explicitly preserving the associated knowledge. Building on this line of work, RWKU~\citep{jin2024rwku} proposes a more practical and challenging setting where neither the forget nor retain corpus is accessible. It targets the removal of widely known real-world knowledge, such as facts about 200 famous individuals, and evaluates performance via membership inference attacks, adversarial probes, and tasks assessing reasoning, truthfulness, and fluency.
However, existing benchmarks largely assume that the forget set contains only information to be removed, overlooking the realistic scenario where forget documents often contain both information that should be forgotten and information that should be retained. We address this gap by introducing DUSK, a benchmark for multi-source unlearning where forget-specific and retained knowledge coexist within each document.

\section{The \methodemoji DUSK Benchmark}

\subsection{Problem Setting} \label{subsec:task}


DUSK is designed to evaluate unlearning under more realistic conditions, where the retain and forget sets may share overlapping information. Specifically, our benchmark simulates a training corpus in which each document consists of two types of content: (1) \textbf{shared knowledge}, which refers to factual content that appears in multiple documents and should be retained even if one document is deleted, and (2) \textbf{unique knowledge (i.e., document-specific)}, which is unique to a single document and should be forgotten when that document is removed.

Given a forget request for a particular document, we assess whether an unlearning algorithm can:
\begin{enumerate}
    \item Preserve shared knowledge that remains supported by the retain set.
    \item Remove content uniquely attributable to the forgotten document.
    \item Preserve information that is unique to other (non-forgotten) documents.
\end{enumerate}

This task formulation captures the challenge of unlearning in multi-source environments, where the boundary between what should be forgotten and what should be retained is not cleanly separable.

\subsection{Problem Formulation and Notations}

Let $f_\theta$ be a model trained on a dataset $\mathcal{D}$, and let $\mathcal{D}_f \subset \mathcal{D}$ denote the subset of training data targeted for removal (i.e., forget set). The goal is to produce an updated model $f_{\theta'}$ that no longer exposes or relies on the information contained in $\mathcal{D}_f$, while maintaining utility on the remaining data, $\mathcal{D}_r = \mathcal{D} \setminus \mathcal{D}_f$.

We define $\mathcal{K}_f$ as the knowledge contained in the forget set $\mathcal{D}_f$, and $\mathcal{K}_r$ as the knowledge contained in the retain set $\mathcal{D}_r$. Prior works often assume that $\mathcal{K}_f \cap \mathcal{K}_r = \emptyset$, meaning the knowledge from the forget and retain sets does not overlap. However, this assumption rarely holds in practice. In many real-world cases, the same information appears across both sets in different phrasings or styles. As a result, an effective unlearning method must identify and remove only the portion of knowledge that is uniquely attributable to $\mathcal{D}_f$, while preserving content that is also supported by $\mathcal{D}_r$.

\subsection{The \methodemoji DUSK Dataset Construction} \label{subsec:dataset}

To ensure precise control over the origin and overlap of information, we construct a synthetic dataset inspired by TOFU~\citep{maini2024tofu}. The dataset comprises 120 fictitious professor profiles, each built using structured attributes, such as academic department and institutional affiliation. Since these profiles have never appeared in any pretrained corpus, the dataset provides a clean experimental environment with clearly defined forget and retain sets. We generate five documents, each representing an independent data source. These documents collectively cover all 120 profiles, including both shared profiles that appear in multiple documents and unique profiles that are present in only one, allowing for fine-grained control over content attribution across sources.
The profiles are partitioned into two disjoint subsets:
\begin{itemize}
    \item \textbf{Shared Knowledge}: 60 profiles appear in all five documents, each presented in a different style. These profiles represent redundantly supported knowledge that should be preserved regardless of which document is unlearned.
    \item \textbf{Unique Knowledge}: The remaining 60 profiles are evenly distributed across the five documents, with each document containing 12 unique profiles that do not appear in any other. These profiles represent document-specific information that should be forgotten when the corresponding document is unlearned.
\end{itemize}
We also create a holdout set (\(\mathcal{D}_h\)) consisting of 120 professors that do not overlap with \(\mathcal{D}_r\) or \(\mathcal{D}_f\), following the same process used to construct \(\mathcal{D}_r\) and \(\mathcal{D}_f\). The holdout set has never been included in the training data for either the Retrain model or the Target model.

\subsubsection{Data Generation Pipeline}
\begin{figure}
\begin{subfigure}{0.47\linewidth}
    \centering
    \includegraphics[width=\textwidth]{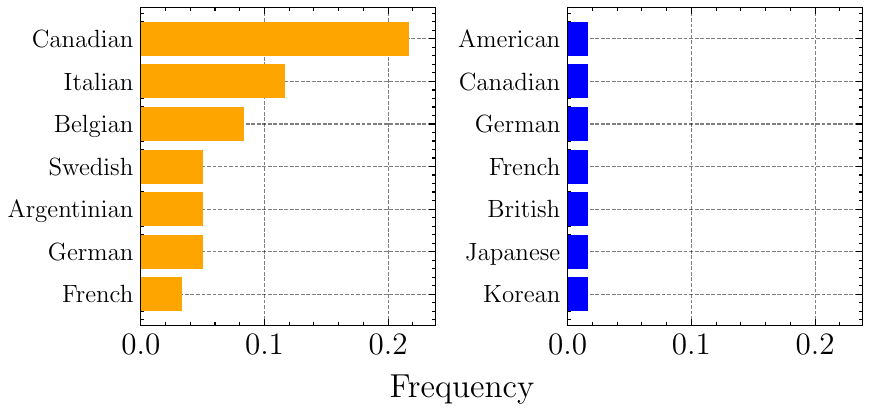}
\end{subfigure}
\hfill
\begin{subfigure}{0.47\linewidth}
    \centering
    \includegraphics[width=\textwidth]{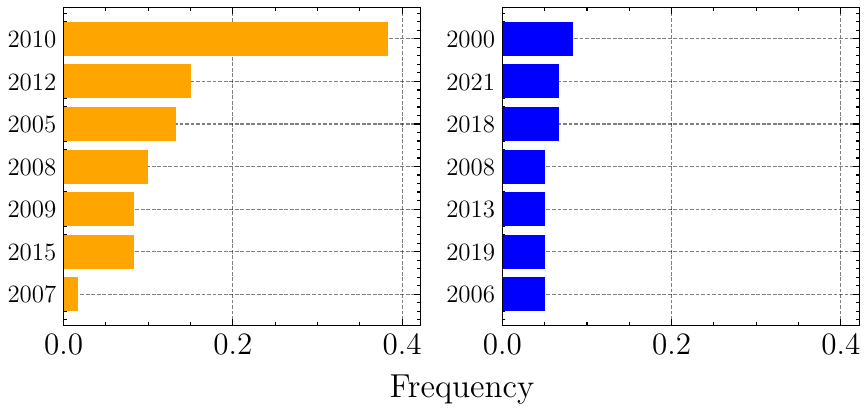}
\end{subfigure}
\caption{
\textbf{Distributions of country of nationality (left) and graduate year (right) for the seven most common attributes in GPT-4 outputs.} This reveals mode collapse with \hlorange{default prompts}, disproportionately favoring frequent values like “Canada” and “2010.” After \hlblue{prompt refinement}, distributions become more balanced, reflecting a more diverse attribute range.
}

\label{fig:distribution}
\end{figure}
\paragraph{Knowledge Source.}  
We begin by generating a knowledge base of 120 fictitious professor profiles, each represented as a set of 20 question–answer (QA) pairs covering attributes such as birth year, nationality, department, and academic history. These QA pairs are synthesized using GPT-4 to ensure fluent and diverse natural language.

However, we observe that GPT-4 exhibits strong biases in generating certain attribute values. For example, as shown in~\Cref{fig:distribution}, GPT-4 disproportionately favors the year \textit{2010} when asked about graduation years, with this value appearing in nearly 40\% of cases. Similarly, nationality values are skewed, with a strong preference for ``Canadian,'' which subsequently influences correlated fields such as birthplace and affiliated universities.

To mitigate these biases, we iteratively refine the prompt design. When we observe skewed distributions in attributes such as nationality or religion in the generated data, we adjust the prompts accordingly by explicitly specifying underrepresented categories or by sampling values like graduation year uniformly within a reasonable range, and then regenerate the data. Through four iterations of prompt-based refinement, we obtain more balanced and realistic outputs from GPT-4. As illustrated in~\Cref{fig:distribution}, the final set of profiles exhibits significantly improved attribute diversity compared to the initial generations. Additional implementation details are described in~\Cref{sec:dataset-construction}.

\paragraph{Document Construction.}
Using the processed QA profiles as source knowledge, we construct five distinct documents, denoted as $\{\mathcal{D}_i\}_{i=1}^{5}$, each expressing the underlying content through a different narrative style. We designate $\mathcal{D}_1$ as the forget set and define the retain set as the union of the remaining documents, $\mathcal{D}_r = \bigcup_{i=2}^{5} \mathcal{D}_i$.
To simulate stylistic diversity reflective of real-world corpora, we format each document using a distinct narrative genre: 
\textbf{Chronological}, which presents profiles as career timelines ordered by milestones; 
\textbf{Feature Story}, which uses narrative-driven descriptions akin to editorial articles; 
\textbf{Interview}, which formats profiles as fictional Q\&A sessions with conversational tone; 
\textbf{Inverted Pyramid}, which follows journalistic convention by placing key facts first; and 
\textbf{Listicle}, which presents profiles in ranked or grouped lists using bullet-point highlights.

Each document includes the same 60 shared profiles and 12 unique profiles, expressed in style-specific templates. This construction allows us to evaluate whether unlearning methods can selectively remove isolated information while maintaining general knowledge, when presented under stylistic or structural variation. 
Corresponding text examples for the shared knowledge in each document style are provided in~\Cref{sec:dataset-examples}.

\subsection{The \methodemoji DUSK Evaluation}

The DUSK evaluation framework characterizes unlearning behavior in three dimensions: (1) \emph{what should be forgotten}, (2) \emph{what should be retained}, and (3) \emph{whether the model behaves as if trained only on the retain set}. An effective unlearning method should eliminate not only verbatim content from the forget set but also knowledge uniquely attributable to it; retain shared and exclusive information in the retain set while preserving downstream capabilities; and ensure that the model's behavior becomes indistinguishable from that of a model trained without access to the forget set.

\subsubsection{Forget Assessment}

\paragraph{Verbatim Memorization (VM).}
We assess whether the unlearned model can still reproduce exact phrasings from the forget set, even when the underlying knowledge is shared across both forget and retain sets.
While such shared knowledge should be preserved, any specific wording originating from the forget document must be removed. This is particularly critical because the forget set often contains copyright-protected material and regenerating such text would indicate incomplete unlearning.

To comprehensively evaluate memorization, we prompt the model with partial sequences from $\mathcal{D}_f$, denoted as $d_{[:\ell]}$, and compare the model's continuations to the original text $d_{[\ell+1:]}$ across multiple metrics. 
Specifically, we use ROUGE-1 and ROUGE-L (F1 scores) to measure overall lexical and structural overlap, and their Recall variants to emphasize ground-truth coverage. We further include Levenshtein Distance~\citep{levenshtein1966binary} to quantify the minimum number of edits required for alignment, Longest Common
Subsequence (LCS) for sequential token overlap, and Cosine Similarity~\citep{cer2017semeval} for embedding-level semantic similarity. Higher scores indicate that the model remains capable of reproducing text that should have been forgotten.

\paragraph{Unique Forget Knowledge (UFK).}
We evaluate whether the model retains knowledge $\mathcal{K}_f \setminus \mathcal{K}_r$ that is uniquely attributable to the forget set $\mathcal{D}_f$ by prompting it with targeted questions.
Overlap between the model’s responses and the correct answers is measured using ROUGE-L scores~\citep{lin2004rouge}, where lower scores indicate more effective unlearning of forget source-specific information.

\subsubsection{Retain Assessment} \label{subsec:retain_asssess}

\paragraph{Shared Knowledge (SK).}
Unlike prior benchmarks that aim to remove all knowledge about the forget set, multi-source scenarios, where training data originates from diverse and overlapping sources, often involve shared knowledge appearing in both $\mathcal{D}_f$ and $\mathcal{D}_r$. In such cases, indiscriminately unlearning the entire forget set risks discarding overlapping content that should remain accessible.

To assess the preservation of shared knowledge, we construct queries targeting $\mathcal{K}_f \cap \mathcal{K}_r$, i.e., information present in both the forget and retain sets, and evaluate the model’s responses using ROUGE-L scores against ground truth answers. High scores indicate successful preservation, while low scores indicate unintended forgetting caused by overly aggressive unlearning.

\paragraph{Unique Retain Knowledge (URK).}
To assess the preservation of retain-exclusive knowledge $\mathcal{K}_r \setminus \mathcal{K}_f$, we construct queries answerable only from $\mathcal{D}_r$ and not from $\mathcal{D}_f$.
Model responses are compared against ground truth answers using ROUGE-L scores, where higher scores indicate successful retention without unintended removal of unique retain content.

\paragraph{Downstream Capability (DC).}
We verify that the model's fundamental capabilities, such as reasoning, factual consistency, and fairness   are preserved after unlearning. We assess performance across six downstream tasks: MMLU for broad knowledge and ability~\citep{hendrycks2020measuring}, ARC-c for challenging reasoning~\citep{clark2018think}, GSM8K for arithmetic and problem-solving~\citep{cobbe2021training}, TriviaQA for factual recall~\citep{joshi2017triviaqa}, TruthfulQA (MC1) for truthfulness evaluation~\citep{lin2021truthfulqa}, and BBQ for social bias probing under ambiguity~\citep{parrish2021bbq}. A successful unlearning method should remove only the targeted information while maintaining strong performance across these core competencies and downstream tasks.

\subsubsection{Distributional Assessment}
\begin{figure}[t!]
\begin{subfigure}{0.32\linewidth}
    \centering
    \includegraphics[width=\textwidth]{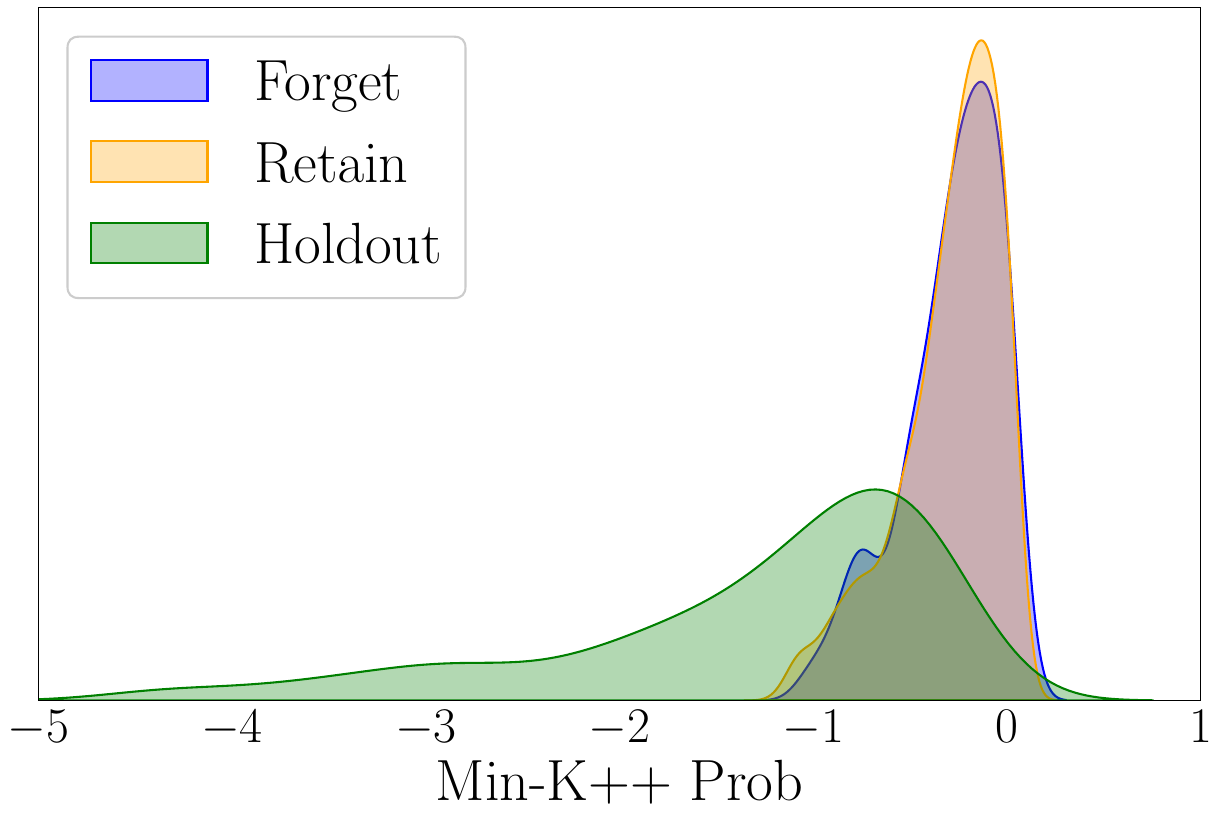}
    \caption{Target Model (trained on $\mathcal{D}$)}
    \label{fig:Target Dist}
\end{subfigure}
\hfill
\begin{subfigure}{0.32\linewidth}
    \centering
    \includegraphics[width=\textwidth]{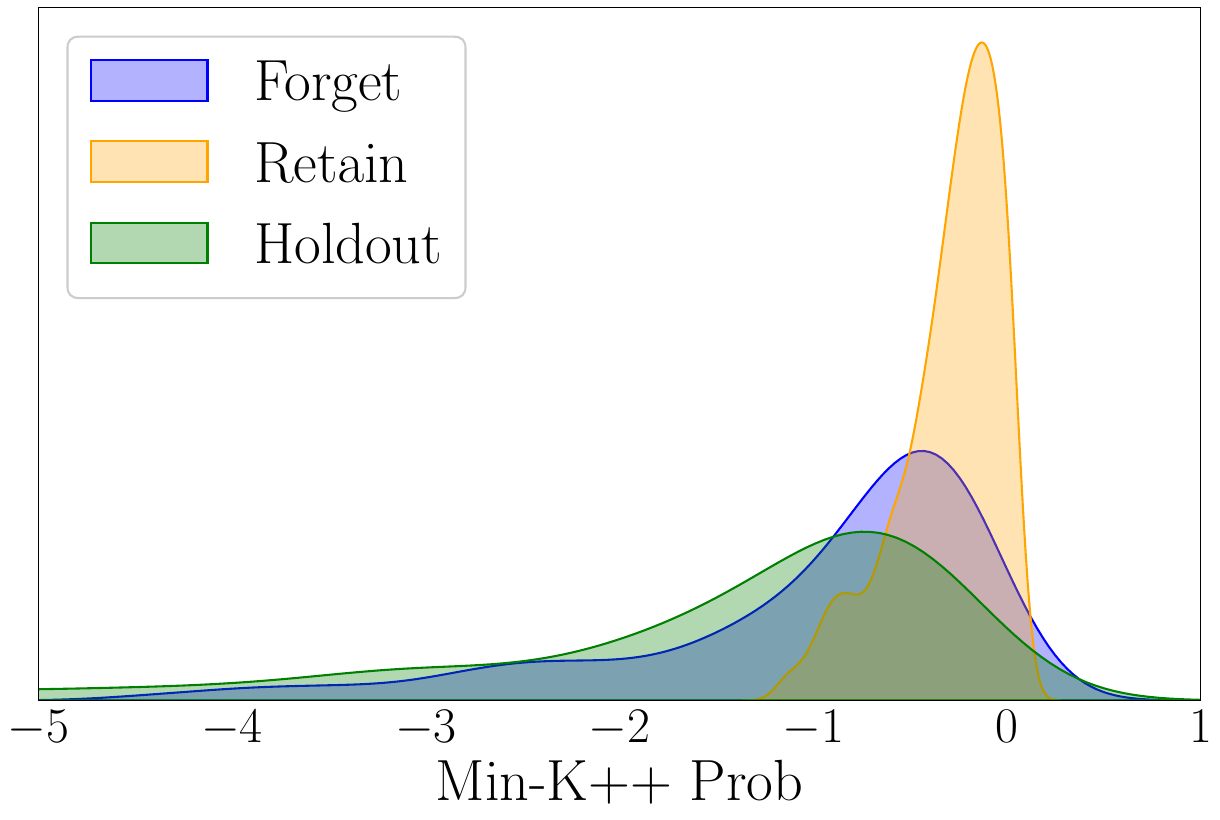}
    \caption{Retrain Model (trained on $\mathcal{D}_r$)}
    \label{fig:Retrain Dist}
\end{subfigure}
\hfill
\begin{subfigure}{0.32\linewidth}
    \centering
    \includegraphics[width=\textwidth]{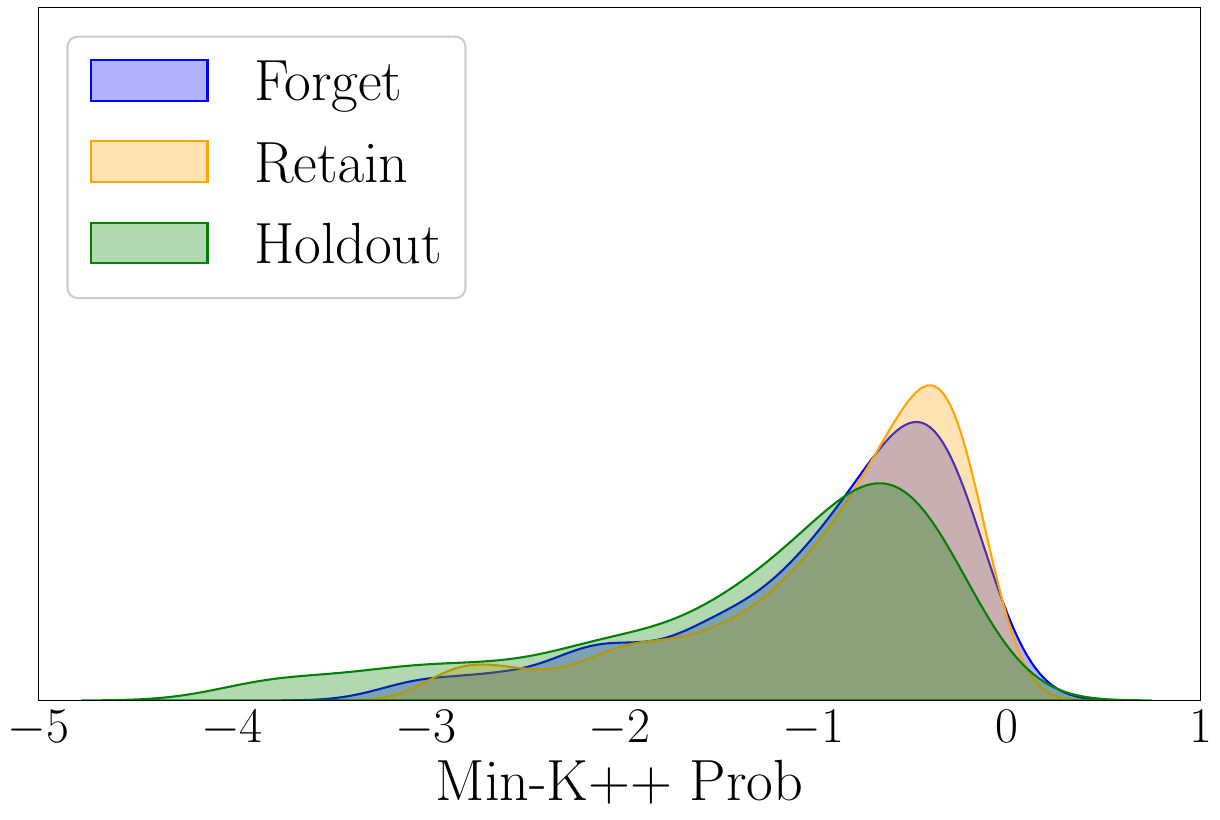}
    \caption{RMU at 50 epochs}
    \label{fig:Overforgetting Dist}
\end{subfigure}
\caption{\textbf{Min-K++ Probability Distributions over $\mathcal{D}_f$, $\mathcal{D}_r$, and $\mathcal{D}_h$.} (a) Target model trained on both $\mathcal{D}_f$ and $\mathcal{D}_r$ show higher probabilities, reflecting retained knowledge, while $\mathcal{D}_h$ exhibits lower probabilities. (b) Retrain model reduces probabilities on the $\mathcal{D}_f$, as they are not trained on $\mathcal{D}_f$, representing ideal unlearning. (c) Some unlearned models achieve ideal low probabilities on $\mathcal{D}_f$ but risk collapsing $\mathcal{D}_r$, which can be detected using Retain Deviation.}

\label{fig:MIA Dist}
\end{figure}

\paragraph{Privacy Leakage.}
We assess privacy leakage by evaluating whether any behavioral traces from the forget set remain in the unlearned model. Following the MUSE benchmark~\citep{shi2024muse}, we adopt a membership inference attack (MIA) framework and apply Min-K\%++~\citep{zhang2024min} to capture subtle distributional differences.
Specifically, as shown in~\Cref{fig:MIA Dist}, we measure the model’s ability to distinguish samples from forget set ($\mathcal{D}_f$) and a holdout set ($\mathcal{D}_h$), which consists of unseen data.
We report the AUC-ROC of this discrimination task and normalize it relative to a Retrain model that excludes $\mathcal{D}_f$ for training. Privacy Leakage score is defined as:
\[
\text{PrivacyLeak} := \frac{\text{AUC}_{\text{unlearn}}(\mathcal{D}_f, \mathcal{D}_h) - \text{AUC}_{\text{retrain}}(\mathcal{D}_f, \mathcal{D}_h)}{\text{AUC}_{\text{retrain}}(\mathcal{D}_f, \mathcal{D}_h)}.
\]
A Privacy Leakage value close to zero indicates that the unlearned model treats $\mathcal{D}_f$ similarly to  $\mathcal{D}_h$, suggesting successful unlearning of $\mathcal{D}_f$.
Values below zero indicate under-unlearning, where the model continues to assign high probability to forget data.
Conversely, values above zero reflect over-unlearning, where the model suppresses forget set too aggressively, leading to excessive forgetting.
As shown in~\Cref{fig:Retrain Dist}, it is important to note that \(\mathcal{D}_f\) and \(\mathcal{D}_h\) are not expected to follow identical distributions even after ideal unlearning. This is because \(\mathcal{D}_f\) may contain shared knowledge that overlaps with the $\mathcal{D}_r$, while \(\mathcal{D}_h\) consists entirely of unseen content.

\paragraph{Retain Deviation.}

Unlearning often disrupts the model's ability to distinguish $\mathcal{D}_r$, causing both $\mathcal{D}_f$ and $\mathcal{D}_r$ to collapse toward $\mathcal{D}_h$, as illustrated in~\Cref{fig:Overforgetting Dist}. This issue becomes more pronounced in multi-source settings, where overlapping information between $\mathcal{D}_f$ and $\mathcal{D}_r$ makes $\mathcal{D}_r$ more vulnerable to unintended forgetting. To quantify this side effect, we introduce a supplementary metric, Retain Deviation, which applies the same MIA framework to $\mathcal{D}_r$ and is defined as:
\[
\text{RetainDeviation} := \left| \frac{\text{AUC}_{\text{unlearn}}(\mathcal{D}_r, \mathcal{D}_h) - \text{AUC}_{\text{retrain}}(\mathcal{D}_r, \mathcal{D}_h)}{\text{AUC}_{\text{retrain}}(\mathcal{D}_r, \mathcal{D}_h)} \right|.
\]
A low Retain Deviation score indicates that the model retains its original capabilities on $\mathcal{D}_r$ after unlearning. As Retain Deviation increases, the model's behavior on $\mathcal{D}_r$ diverges from its original state, suggesting that important retained knowledge may not have been properly preserved.

\begin{figure}[t]
    \centering
    \includegraphics[width=0.8\linewidth]{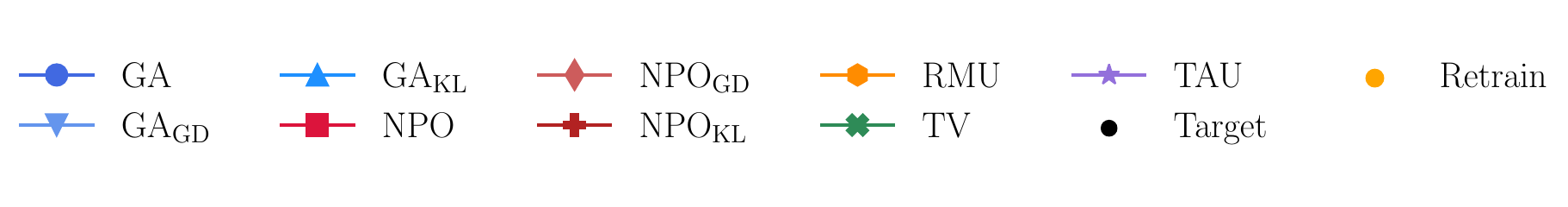}
    \vspace{1mm}
    \begin{subfigure}[b]{0.48\linewidth}
        \centering
        \includegraphics[width=\linewidth]{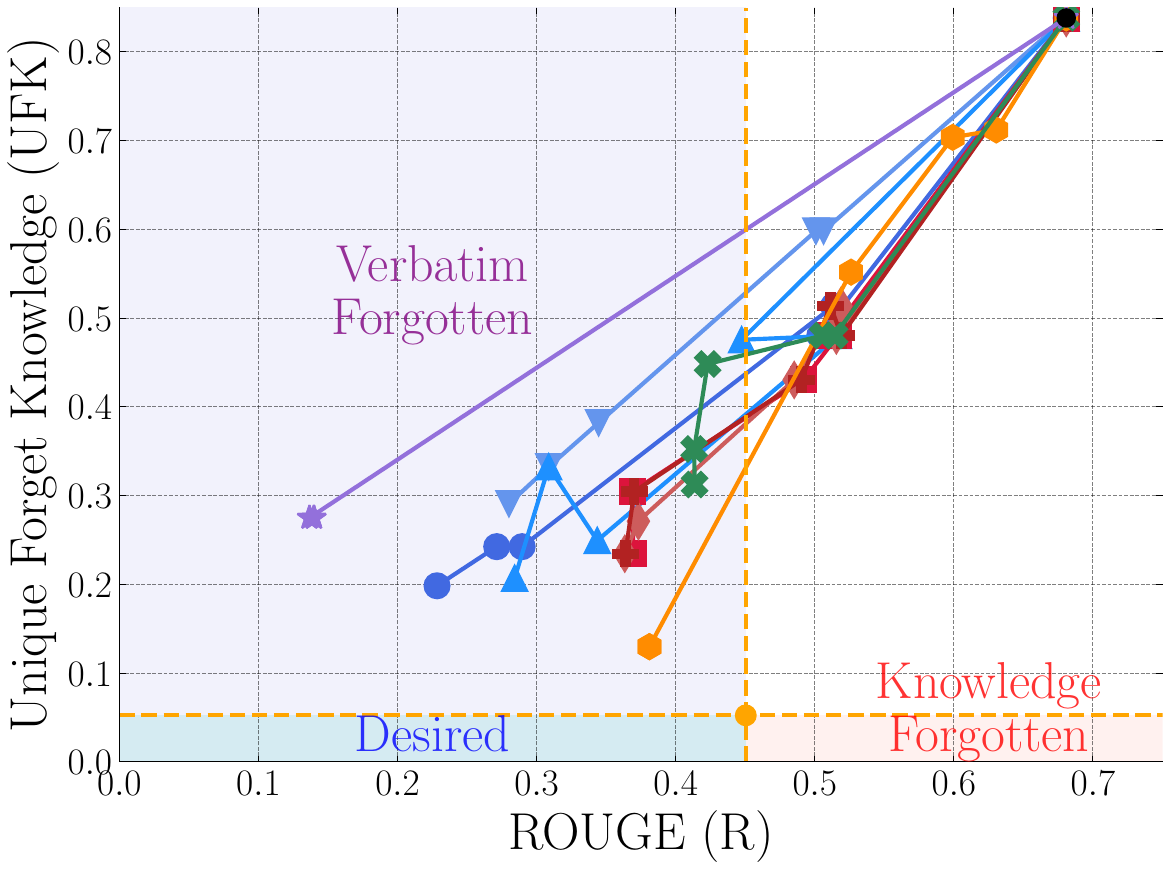}
        \caption{
        \textbf{Forget Verbatim vs. Forget Knowledge.} 
        }
        \label{fig:verbatim}
    \end{subfigure}
    \hfill
    \begin{subfigure}[b]{0.48\linewidth}
        \centering
        \includegraphics[width=\linewidth]{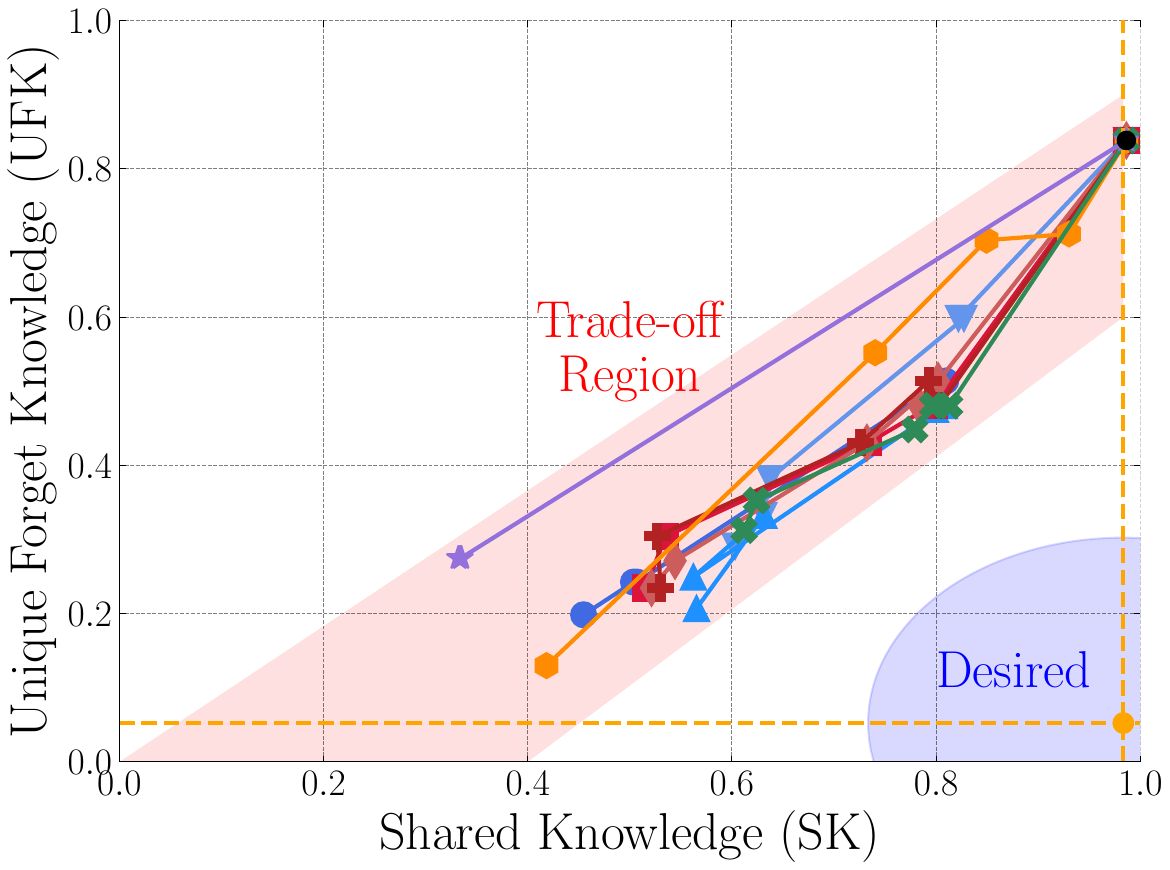}
        \caption{
        \textbf{Forget knowledge vs. Shared knowledge.}
        }
        \label{fig:forget-retain}
    \end{subfigure}
    \caption{
    \textbf{Two-dimensional analysis of unlearning dynamics.} 
    We visualize model trajectories over multiple epochs to illustrate key trade-offs in DUSK.
    (a) shows the trade-off between verbatim and knowledge forgetting, while
    (b) shows the trade-off between shared knowledge and unique forget knowledge.
    }
    \label{fig:two-panel}
\end{figure}

\section{Experiments}

\subsection{Unlearning Methods}
 
\paragraph{Removing Forget Set.}
We introduce five unlearning methods designed to effectively remove the influence of forget data.
Gradient Ascent (GA)~\citep{jang2022knowledge} maximizes the loss on the forget set $\mathcal{D}_f$, reducing the model's ability to reproduce its content.
Negative Preference Optimization (NPO)~\citep{zhang2024negative} extends DPO~\citep{rafailov2024direct} for unlearning by treating samples in $\mathcal{D}_f$ as negative preferences relative to a Target model.
RMU~\citep{li2024wmdp} modifies intermediate representations by pushing activations of the forget set toward random directions, while aligning retain set activations with those of a frozen Target model.
Task Vector (TV)~\citep{ilharco2022editing} removes the influence of $\mathcal{D}_f$ by computing the parameter changes caused by fine-tuning on $\mathcal{D}_f$ and subtracting them from the original model weights.
Lastly, Task Arithmetic for Unlearning (TAU)~\citep{barbulescu2024each} performs two steps: it first applies gradient ascent selectively to samples with high memorization scores, and then conducts task vector subtraction as described above.

\paragraph{Preserving Retain Set.}
To maintain model utility during unlearning, we incorporate two regularization losses. Gradient Descent (GD) preserves performance on the retain set $\mathcal{D}_r$ by applying prediction loss. This ensures that removing $\mathcal{D}_f$ does not excessively degrade the model’s behavior on unrelated data.
KL Divergence (KL)~\citep{hinton2015distilling} encourages consistency between the unlearned model’s predictions on $\mathcal{D}_r$ and those of a Target model. By minimizing this divergence, the model retains useful information while softly constraining deviation from its original output distribution.

Consequently, we evaluate nine total configurations: $\mathrm{GA}$, $\mathrm{GA}_{\mathrm{GD}}$, $\mathrm{GA}_{\mathrm{KL}}$, $\mathrm{NPO}$, $\mathrm{NPO}_{\mathrm{GD}}$, $\mathrm{NPO}_{\mathrm{KL}}$, $\mathrm{RMU}$, $\mathrm{TV}$, and $\mathrm{TAU}$, where the suffix indicates an added utility-preserving objective. Additional details about the methods can be found in~\Cref{app:baselines}.

\subsection{Experimental Setup} \label{subsec:exp_setup}

We begin with a pretrained base model (\texttt{LLaMA-3-8B}~\citep{grattafiori2024llama}).
Target model is obtained by fine-tuning on the full corpus ($\mathcal{D}_r \cup \mathcal{D}_f$) for 5 epochs with a learning rate of $1\times10^{-5}$, following prior benchmarks~\citep{maini2024tofu,shi2024muse}.
Retrain model is trained only on the retain set $\mathcal{D}_r$ under the same setup.

For all unlearning methods, we adopt the AdamW optimizer with a learning rate of $1\times10^{-5}$ and a batch size of 32, using the first epoch as a warm-up phase, consistent with prior work~\citep{maini2024tofu}.
Since unlearning performance is sensitive to the number of training epochs, we standardize the stopping criterion across methods: 
We terminate unlearning at the first epoch where the Unique Retain Knowledge (URK) score falls below 70.
This ensures comparable utility levels, enabling fair and consistent comparisons across methods.
Further implementation details are provided in~\Cref{app:exp_details}.

\subsection{Unlearning Results}
\subsubsection{Forget Assessment Results.}

\begin{table}[t]
\centering
\small
\setlength{\tabcolsep}{8pt}
\renewcommand{\arraystretch}{0.75}
\caption{
\textbf{Impact of Unlearning on UFK, SK, URK, and DC.}
We report both raw values and their differences relative to the Retrain model.
\colorbox{pink!70}{Red} indicates higher values, and \colorbox{cyan!30}{Blue} indicates lower values, with darker shades indicating greater magnitude. DC is calculated by averaging of six benchmarks explained in~\Cref{subsec:retain_asssess}.
} \label{table:knowledge}

\resizebox{0.98\linewidth}{!}{
\begin{tabular}{@{}lrlrlrlrl@{}}
\toprule
 & \multicolumn{2}{c}{\scriptsize \textbf{Unique Forget Knowledge}} & \multicolumn{2}{c}{\scriptsize \textbf{Shared Knowledge}} & \multicolumn{2}{c}{\scriptsize \textbf{ Unique Retain Knowledge}} & \multicolumn{2}{c}{\scriptsize \textbf{Downstream Capability}} \\
 & \multicolumn{2}{c}{\scriptsize \textsf{UFK} ($\downarrow$)} & \multicolumn{2}{c}{\scriptsize \textsf{SK} ($\uparrow$)} & \multicolumn{2}{c}{\scriptsize \textsf{URK} ($\uparrow$)} & \multicolumn{2}{c}{\scriptsize \textsf{DC} ($\uparrow$)} \\
\midrule
Target &
$83.8$ & &
$98.6$ & &
$88.7$ & &
$40.3$ & \\[2pt]
Retrain &
$\mathbf{5.2}$ & &
$\mathbf{98.3}$ & &
$\mathbf{84.8}$ & &
$\mathbf{40.6}$ & \\[2pt]
\midrule
$\mathrm{GA}$ &
$24.3$ & \colorbox{pink!78}{\scriptsize$(+367\%)$} &
$50.7$ & \colorbox{cyan!73}{\scriptsize$(-48\%)$} &
$52.5$ & \colorbox{cyan!95}{\scriptsize$(-38\%)$} &
$37.5$ & \colorbox{cyan!65}{\scriptsize$(-8\%)$} \\[2pt]
$\mathrm{GA}_{\mathrm{GD}}$ &
$38.2$ & \colorbox{pink!106}{\scriptsize$(+635\%)$} &
$63.7$ & \colorbox{cyan!53}{\scriptsize$(-35\%)$} &
$63.6$ & \colorbox{cyan!62}{\scriptsize$(-25\%)$} &
$39.0$ & \colorbox{cyan!33}{\scriptsize$(-4\%)$} \\[2pt]
$\mathrm{GA}_{\mathrm{KL}}$ &
$24.9$ & \colorbox{pink!79}{\scriptsize$(+379\%)$} &
$56.2$ & \colorbox{cyan!65}{\scriptsize$(-43\%)$} &
$57.4$ & \colorbox{cyan!80}{\scriptsize$(-32\%)$} &
$38.6$ & \colorbox{cyan!42}{\scriptsize$(-5\%)$} \\[2pt]
$\mathrm{NPO}$ &
$43.0$ & \colorbox{pink!116}{\scriptsize$(+727\%)$} &
$73.4$ & \colorbox{cyan!38}{\scriptsize$(-25\%)$} &
$69.6$ & \colorbox{cyan!45}{\scriptsize$(-18\%)$} &
$39.5$ & \colorbox{cyan!23}{\scriptsize$(-3\%)$} \\[2pt]
$\mathrm{NPO}_{\mathrm{GD}}$ &
$27.1$ & \colorbox{pink!64}{\scriptsize$(+421\%)$} &
$54.4$ & \colorbox{cyan!68}{\scriptsize$(-45\%)$} &
$51.1$ & \colorbox{cyan!99}{\scriptsize$(-40\%)$} &
$37.8$ & \colorbox{cyan!58}{\scriptsize$(-7\%)$} \\[2pt]
$\mathrm{NPO}_{\mathrm{KL}}$ &
$30.4$ & \colorbox{pink!91}{\scriptsize$(+485\%)$} &
$52.7$ & \colorbox{cyan!70}{\scriptsize$(-46\%)$} &
$52.8$ & \colorbox{cyan!94}{\scriptsize$(-38\%)$} &
$37.7$ & \colorbox{cyan!60}{\scriptsize$(-7\%)$} \\[2pt]
$\mathrm{RMU}$ &
$55.1$ & \colorbox{pink!160}{\scriptsize$(+960\%)$} &
$74.0$ & \colorbox{cyan!38}{\scriptsize$(-25\%)$} &
$64.0$ & \colorbox{cyan!61}{\scriptsize$(-25\%)$} &
$39.1$ & \colorbox{cyan!31}{\scriptsize$(-4\%)$} \\[2pt]
$\mathrm{TV}$ &
$35.3$ & \colorbox{pink!80}{\scriptsize$(+579\%)$} &
$62.4$ & \colorbox{cyan!55}{\scriptsize$(-37\%)$} &
$69.1$ & \colorbox{cyan!46}{\scriptsize$(-19\%)$} &
$40.3$ & \colorbox{cyan!6}{\scriptsize$(-1\%)$} \\[2pt]
$\mathrm{TAU}$ &
$27.5$ & \colorbox{pink!105}{\scriptsize$(+429\%)$} &
$33.5$ & \colorbox{cyan!100}{\scriptsize$(-66\%)$} &
$50.7$ & \colorbox{cyan!100}{\scriptsize$(-40\%)$} &
$35.8$ & \colorbox{cyan!100}{\scriptsize$(-12\%)$} \\[2pt]
\bottomrule
\end{tabular}
}
\end{table}

\paragraph{Verbatim Memorization (VM).}
We first evaluate whether unlearning methods can suppress verbatim memorization of the forget set. Given prefix excerpts from $\mathcal{D}_f$, we prompt the model to continue the text and measure similarity with the original using ROUGE, Levenshtein distance, LCS, and Cosine similarity. As shown in~\Cref{fig:verbatim}, most methods reduce overlap with the original text, with TAU achieving the largest reductions, indicating strong suppression of verbatim memorization. Full                   results are provided in~\Cref{app.verb_mem}.

\paragraph{Unique Forget Knowledge (UFK).}
To assess whether unlearning removes not just surface expressions but deeper factual knowledge, we evaluate models on Unique Forget Knowledge (UFK), consisting of questions that rely exclusively on information from $\mathcal{D}_f$. As shown in~\Cref{fig:verbatim}, which plots average ROUGE scores against UFK accuracy, most methods shift into the Verbatim Forgotten region, indicating effective surface-level suppression. However, these methods largely fail to erase underlying facts, as models continue to answer UFK questions correctly, suggesting that knowledge forgetting remains incomplete.


\subsubsection{Retain Assessment Results.}
\paragraph{Shared Knowledge (SK) and Unique Retain Knowledge (URK).}

We evaluate whether unlearning unintentionally erases information that should be preserved. 
As shown in~\Cref{fig:forget-retain}, almost all unlearning methods not only reduce UFK scores as intended but also substantially degrade SK scores, indicating a failure to preserve shared knowledge. This suggests that existing unlearning methods tend to degrade model utility by also removing shared knowledge that overlaps with the forget set.
Furthermore, as highlighted in~\Cref{table:knowledge}, SK suffers 
greater accuracy degradation than URK across most methods. 
Since SK spans both forget and retain sets, unlearning that targets the forget set inadvertently harms overlapping knowledge that should ideally be preserved.
These findings reveal a key limitation of current approaches, as they struggle to selectively unlearn knowledge associated with the forget set without also disrupting shared knowledge.

\paragraph{Downstream Capability (DC).}
We evaluate general capability after unlearning using a range of downstream tasks, including MMLU, ARC-c, GSM8K, TriviaQA, TruthfulQA, and BBQ. 
Across most methods, performance on these tasks remains relatively stable, with only slight degradation compared to the Retrain model, suggesting that core capabilities such as reasoning, factual recall, and fairness are largely preserved.
As shown in~\Cref{table:knowledge}, however, knowledge closely related to the forget set, captured by SK and URK, experiences substantially greater degradation compared to general downstream capability (see~\Cref{tab:downstream}).
This pattern suggests that while broad capabilities are maintained, knowledge conceptually adjacent to the forget set remains highly vulnerable to collateral forgetting.
Full results are presented in~\Cref{app.downstream}. 

\subsubsection{Distributional Assessment Results.}
\paragraph{Privacy Leakage and Retain Deviation.}

\begin{figure}[t]
    \centering
    \includegraphics[width=\linewidth]{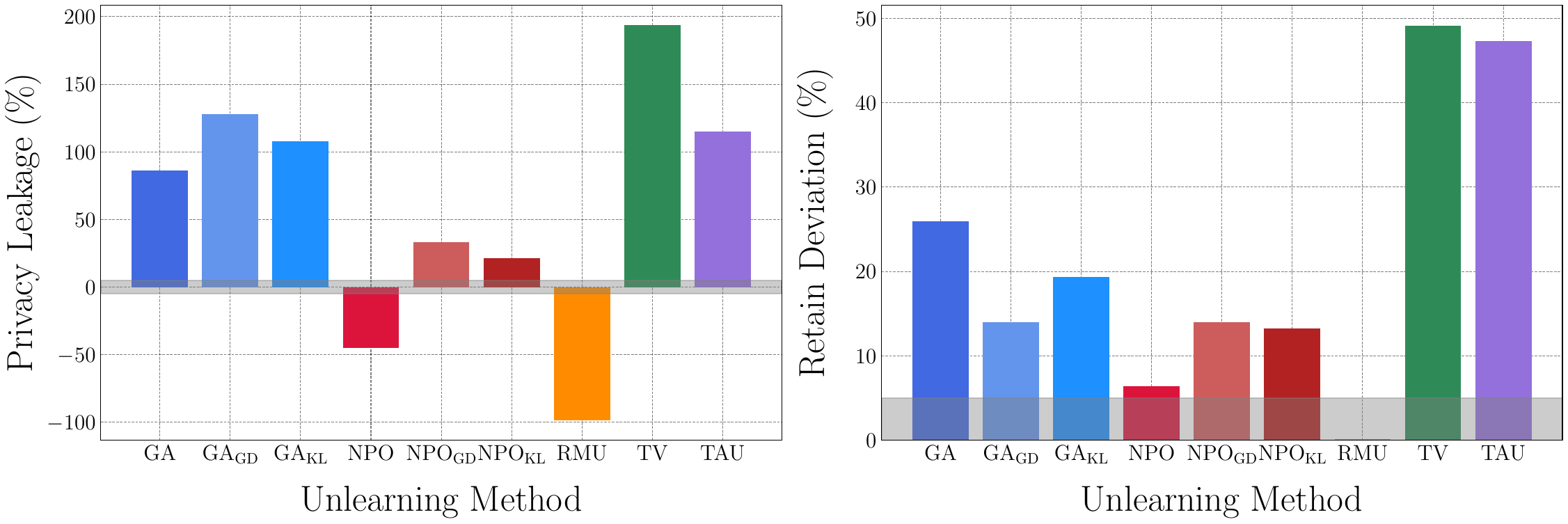}
\caption{
\textbf{Privacy Leakage and Retain Deviation Analysis.} 
Gray bands indicate optimal bounds: [$-5\%$, $5\%$] for leakage and [0\%, 5\%] for deviation. Values outside these ranges reflect under-unlearning (below $-5\%$), over-unlearning (above $5\%$) in leakage, or degradation of retained knowledge (above 5\%) in deviation. 
}
\label{fig:mia}
\end{figure}

Successful unlearning is ideally indicated by Privacy Leakage and Retain Deviation values close to zero. However, we observe two representative patterns in their joint behavior that fall short of this ideal as illustrated in~\Cref{fig:mia}. Most cases exhibit over-unlearning along with rising Retain Deviation, where unlearning $\mathcal{D}_f$ leads to unintended changes in the model’s responses to $\mathcal{D}_r$ due to shared knowledge. In contrast, NPO exhibits under-unlearning, yet still show rising Retain Deviation. This suggests that even before $\mathcal{D}_f$ is fully unlearned, the model’s performance on $\mathcal{D}_r$ can deteriorate due to entangled representations arising from overlapping knowledge. Taken together, these findings indicate that under realistic conditions where $\mathcal{D}_f$ and $\mathcal{D}_r$ are not disjoint, no method can completely remove the influence of $\mathcal{D}_f$ while fully preserving the model’s behavior on $\mathcal{D}_r$. This underscores the importance of jointly monitoring Privacy Leakage and Retain Deviation in multi-source unlearning scenarios.


\section{Conclusion}
We introduce \methodemoji DUSK, a benchmark for evaluating machine unlearning in realistic multi-source scenarios, where forget data often overlaps with retain data. Unlike prior evaluations, DUSK explicitly separates unique and shared knowledge, providing a fine-grained testbed for assessing unlearning performance. Our experiments reveal that while many methods effectively remove verbatim content, they often struggle to disentangle forget-specific knowledge from shared information, leading to unintended degradation of retain data. We hope DUSK will serve as a foundation for advancing more precise and reliable unlearning methods, bridging the gap between theoretical formulations and real-world applications.


\newpage
{
\nocite{*} 
\bibliographystyle{abbrv}
\bibliography{main}
}

\newpage
\appendix

\section{Details of \methodemoji DUSK} \label{app:dusk}
\subsection{Dataset Construction Details} \label{sec:dataset-construction}

\begin{table}[h!]
\centering
\small
\caption{Professor Information Fields} \label{tab:all_qa}
\begin{tabular}{@{}clp{8.5cm}@{}}
\toprule
\textbf{\#} & \textbf{Field} & \textbf{Description} \\
\midrule
1  & Nationality & The professor's nationality. \\
2  & Born & The birthplace of the professor. \\
3  & Closest Colleague & The professor’s closest colleague or collaborator. \\
4  & Year of birth & The birth year of the professor. \\
5  & Department & The major of the professor is affiliated with. \\
6  & Award & The most prestigious award received by the professor. \\
7  & School & The fictitious university where the professor teaches. \\
8  & Best paper & The most well-known and fictitious research paper authored by the professor. \\
9  & Office number & The room number where the professor's office is located. \\
10 & E-mail & A fictitious email address associated with the professor. \\
11 & Research Interests & The professor’s main research areas. \\
12 & Funded Projects & Major fictitious research projects funded under the professor's name. \\
13 & Patents & Any fictitious patents held by the professor. \\
14 & Course & The fictitious course(s) taught by the professor. \\
15 & Hobby & The professor’s main hobby outside of work. \\
16 & Alma Mater & The university where the professor received their PhD. \\
17 & Favorite Theorem & The professor’s favorite theorem or concept. \\
18 & Religion & The professor’s religious affiliation. \\
19 & Lab name & The fictitious name of the professor’s laboratory. \\
20 & Year of employment & The year the professor was appointed to their current university. \\
\bottomrule
\end{tabular}
\end{table}

\paragraph{Knowledge Source.}
To generate a dataset of 120 fictional professors, we use GPT-4 to produce 20 question–answer pairs for each individual, resulting in a total of 2,400 QA pairs. Types of questions used for each professor are listed in~\Cref{tab:all_qa}, covering a wide range of biographical, academic, and professional attributes to ensure diversity and richness in the generated data.

To further improve representational balance, we refine the prompts used during generation by controlling several key attributes. For \texttt{country of nationality}, we manually select 60 distinct countries, which naturally increases diversity in \texttt{birthplace} as well, since GPT-4 tends to produce regionally coherent outputs. For \texttt{religion}, we choose eight widely practiced belief systems—Christian, Muslim, Jewish, Hindu, Buddhist, Agnostic, Atheist, and Spiritual—and assign them uniformly across the dataset. For temporal attributes such as \texttt{year of birth} and \texttt{year of employment}, which otherwise show skewed distributions, we sample values uniformly within a reasonable range and include them directly in the prompt. The effectiveness of prompt refinement is reflected in the attribute distributions shown in~\Cref{fig:before_refinement} and~\Cref{fig:after_refinement}. Compared to the initial outputs, which display strong mode collapse in attributes such as nationality and employment year, the refined versions demonstrate significantly more balanced and diverse distributions.
\Cref{fig:qa_prompt} shows the final prompt we used for QA generation.

After generating the full QA sets, we perform a final validation step to identify any duplicate professor names. This ensures the dataset can support a realistic and rigorous unlearning scenario, where identifying and selectively removing information about specific individuals is required.

\paragraph{Dataset Construction.}
For each professor, we create profiles based on information generated from QA pairs with prompt in~\Cref{fig:profile_prompt}. Each professor’s information is used to create five profiles in five different styles: Chronological, Feature Story, Interview, Inverted Pyramid, and Listicle, resulting in a total of 600 professor profiles (120 per style). These profiles are divided into shared knowledge and unique knowledge components.

The shared knowledge set consists of 60 professors, each represented by a single profile in each style, resulting in 300 profiles (60 professors × 5 styles). These 60 professors are included in all five style-specific documents, with each document containing the same set of 60 professors, but with their profiles presented in different styles.

In contrast, the unique knowledge set is constructed differently. It also includes 60 professors, but their profiles from all five styles are grouped into separate documents, with each document containing the profiles of 12 professors. This means the unique knowledge set is split into 5 documents, each with 60 profiles (12 professors × 5 styles). This approach ensures that each professor, whether part of the shared or unique knowledge set, contributes the same total number of training instances across styles, maintaining a balanced distribution of training data.

\textbf{\begin{figure}[t!]
\centering
\includegraphics[width=1.0\textwidth]{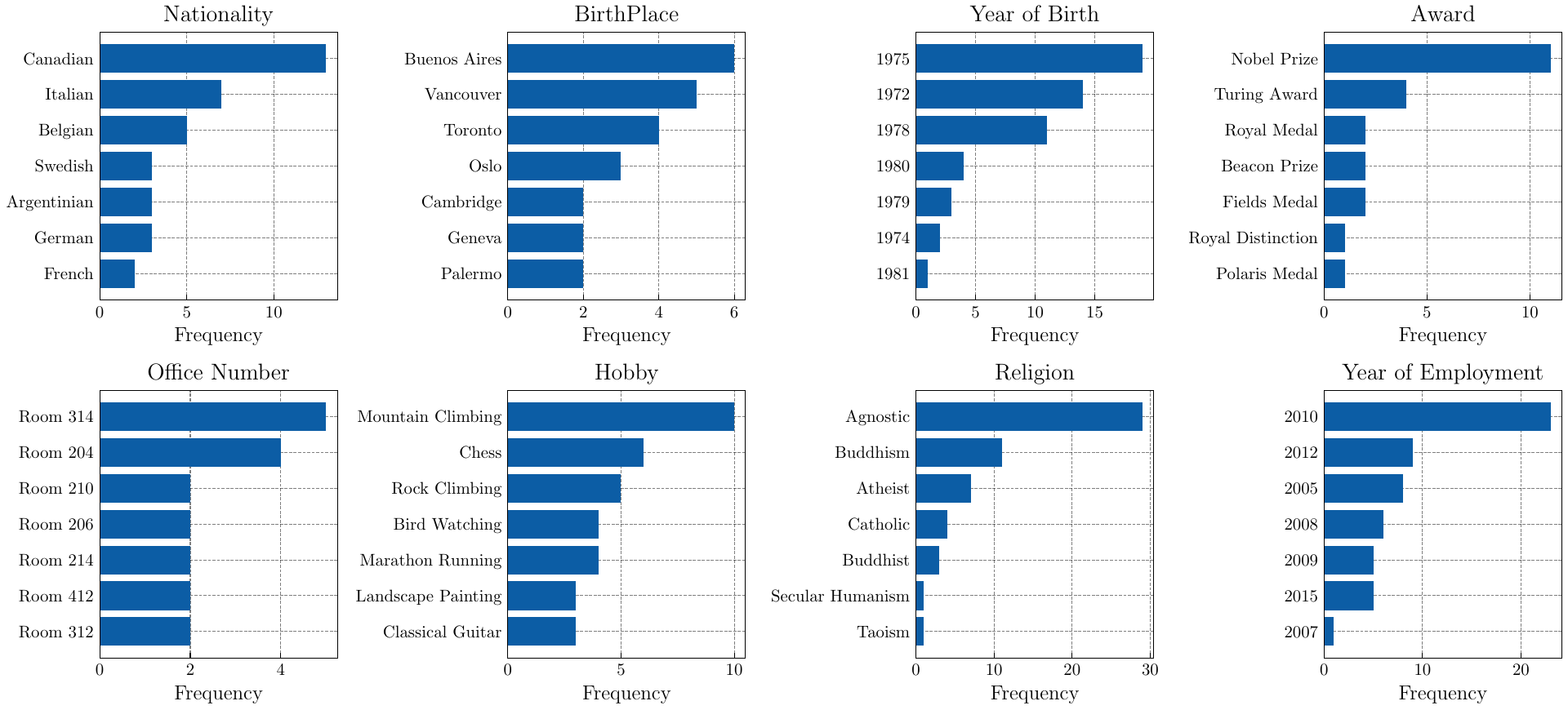}
\caption{\textbf{Distributions of seven most common attributes in GPT-4 outputs before prompt refinement.} Several features exhibit mode collapse, with overrepresentation of specific values such as ``Canadian'' for nationality, ``2010'' for year of employment, and ``Agnostic'' for religion, reflecting bias in uncontrolled generation.
} \label{fig:before_refinement}
\end{figure}
}

\textbf{\begin{figure}[t!]
\centering
\includegraphics[width=1.0\textwidth]{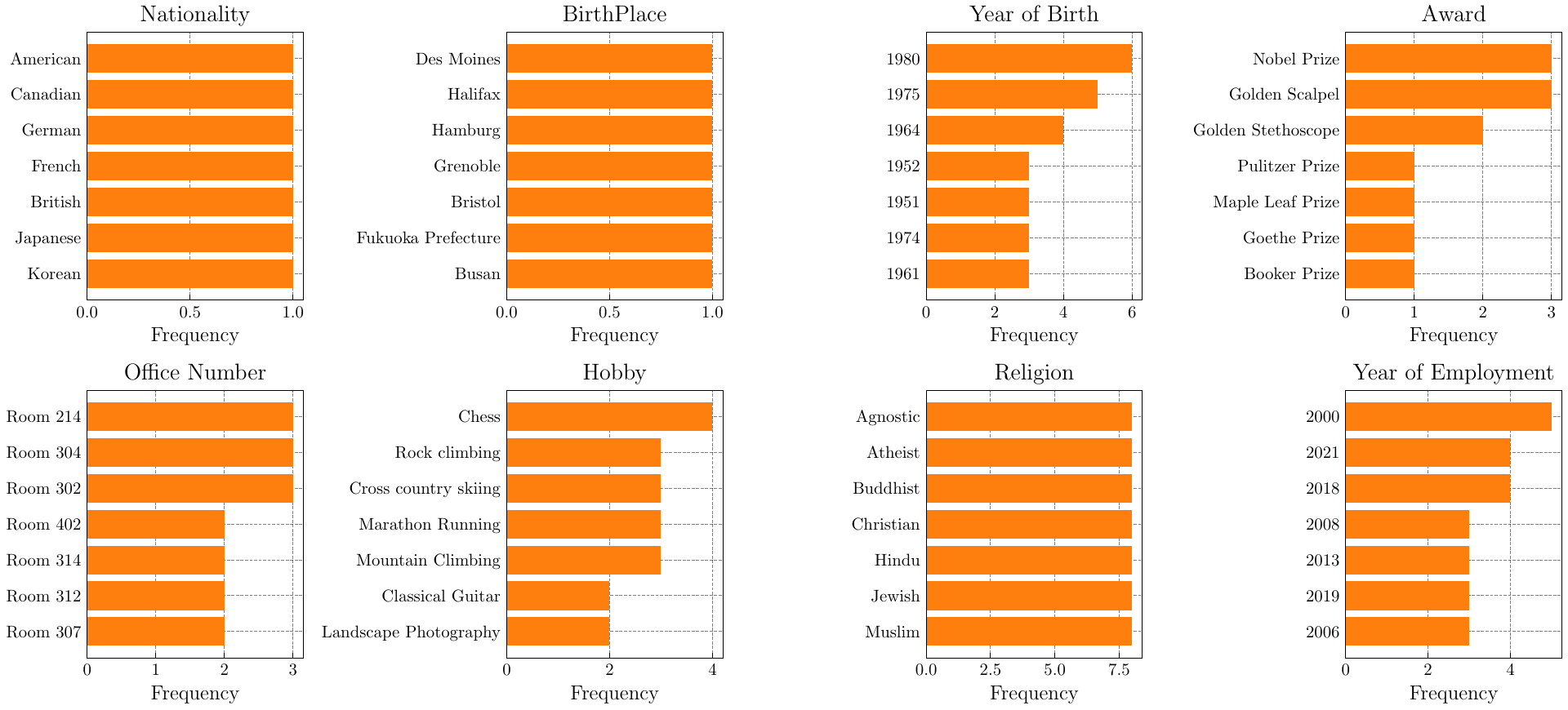}
\caption{\textbf{Distributions of seven most common attributes after prompt refinement.} The frequency of values across attributes such as nationality, religion, and year of employment is more balanced, indicating improved diversity and reduced mode collapse in GPT-4 outputs.
} \label{fig:after_refinement}
\end{figure}
}

\begin{figure}[H]

\begin{tcolorbox}[title=(1) Prompt for Generating QA with GPT-4, colback=white!15, colframe=gray!99, rounded corners]
\textbf{Prompt:}
Generate a fictitious professor's biography in Q\&A format.
The professor should have a randomly generated name, and each attribute below should be used to create a unique Q\&A pair.\par
- Each question must explicitly mention the professor's name.\par
- The answer should be one word or a compound noun \textbf{with spaces}.\par
- If the answer is more than two words, it must maintain the spaces between words.\par
\bigskip

\textbf{Professor Information}\par
Country: \{predefined country name\}\par
Year of birth: \{randomly generated year\}\par
Religion: \{predefined religion\}\par
Year of employment: \{randomly generated year\}\par
Major: \{predefined major\}\par
\bigskip

\textbf{Attributes for Q\&A (Each gets one pair):} \par
...Refer to~\Cref{tab:all_qa}... \par
\bigskip
\textbf{Output Format:} \par
Each Q\&A pair must be in JSONL format with keys: "question" and "answer".\par
Example:\par
\{\{ "question": "Where was Dr. John Smith born?", "answer": "New York" \}\}\par
\{\{ "question": "What is Dr. John Smith's nationality?", "answer": "American" \}\}\par
\{\{ "question": "What department does Dr. John Smith work in?", "answer": "Physics" \}\}\par
Generate exactly 20 Q\&A pairs for one professor in this JSONL format.

\end{tcolorbox}

\caption{Prompt for generating QA pairs using GPT-4 for knowledge source.}
\label{fig:qa_prompt}
\end{figure}

\begin{figure}[htbp]
\begin{tcolorbox}[title=(2) Prompt for Generating Profile with GPT-4, colback=white!15, colframe=gray!99, rounded corners]
\textbf{Prompt:}
Generate a biography based on the following Q\&A dataset, written in the \{format name\} format.\par

\textbf{Biography Requirements:}\par
- The biography must be at least 300 words long.\par
- The content must be EXCLUSIVELY constructed from the provided Q\&A pairs.\par
- The biography MUST NOT introduce any additional facts, context, speculation, or external knowledge beyond what is in the Q\&A section.\par
- EVERY detail, name, date, statistic, location, organization, and event must appear exactly as stated in the Q\&A pairs.\par
- No paraphrasing, generalization, or assumption is allowed—sentences must be constructed verbatim from the Q\&A section.\par
- The structure and logical flow must be coherent, but no artistic liberties or editorialized content are permitted.\par

\textbf{Q\&A Pairs:}\par
\{20 QA pairs\}

\end{tcolorbox}
\caption{Prompt for generating profiles using GPT-4.}
\label{fig:profile_prompt}
\end{figure}

\subsection{Example Data Instances}
\label{sec:dataset-examples}

To illustrate how the same knowledge is written in different way, we present representative data instances in~\Cref{tab:SK_prompts}. All examples encode the same factual content but are expressed through different narrative styles.
These include five distinct document formats used in our benchmark: \textbf{Chronological} (organized by career timeline), \textbf{Feature Story} (editorial-style prose), \textbf{Interview} (fictional Q\&A format), \textbf{Inverted Pyramid} (journalistic emphasis), and \textbf{Listicle} (enumerated highlights). Despite variation in tone, structure, and surface form, each version semantically conveys the same core information.
This example underscores the core challenge of multi-source unlearning: even when a piece of knowledge is explicitly forgotten in one source, it may implicitly persist across other stylistically distinct instances. Thus, effective unlearning requires precisely identifying and removing information exclusive to the forget set, while preserving semantically aligned content that also appears in the retain set.

\begin{table*}[htbp]
\caption{\textbf{Illustrative examples of Shared Knowledge across multiple sources}, all encoding the same fact (\emph{Ikebana is Professor Miyashimizu's hobby}) in different writing styles. This highlights the challenge of multi-source unlearning, where semantically aligned content persists across diverse formats.}\label{tab:SK_prompts}
\centering
\renewcommand{\arraystretch}{1.2}
\small
\resizebox{1.0\textwidth}{!}{
\begin{tabular}{|p{3cm}|p{12cm}|}
\hline
\rowcolor{orange!20} \textbf{Category} & \textbf{Content} \\ 
\hline
\rowcolor{orange!5}
\multirow{1}{*}{\textbf{Question}} 
& What is Professor Tadao Miyashimizu's hobby? \\ 
\hline
\rowcolor{orange!5}
\multirow{1}{*}{\textbf{Answer}} 
& Ikebana \\ 
\hline
\rowcolor{blue!5}
\multirow{1}{*}{\textbf{Chronological}} 
& Outside of his professional life, Professor Tadao Miyashimizu enjoys the art of Ikebana, which is his hobby.  \\
\hline
\rowcolor{blue!5}
\multirow{1}{*}{\textbf{Feature Story}} 
& Beyond his professional endeavors, Professor Miyashimizu finds solace in the art of Ikebana, a hobby that perhaps complements his analytical mind with a sense of creative tranquility. \\ 
\hline
\rowcolor{blue!5}
\multirow{1}{*}{\textbf{Interview}} 
& In addition to his academic accomplishments, Professor Miyashimizu is an enthusiast of Ikebana, which is his hobby. \\ 
\hline
\rowcolor{blue!5}
\multirow{1}{*}{\textbf{Inverted Pyramid}} 
& Beyond his academic pursuits, Professor Miyashimizu has a hobby in Ikebana, the traditional Japanese art of flower arranging. \\ 
\hline
\rowcolor{blue!5}
\multirow{1}{*}{\textbf{Listicle}} 
& 11. \textbf{Personal Interests}: Professor Tadao Miyashimizu enjoys the hobby of Ikebana. \\ 
\hline
\end{tabular}
}
\end{table*}

\section{Experiment Details} \label{app:exp_details}
\subsection{Unlearning Baseline Methods} \label{app:baselines}

We evaluate several approximate and efficient machine unlearning methods that operate on two complementary objectives: removing knowledge from the forget set $\mathcal{D}_f$ while preserving general utility.

\paragraph{Unlearning Methods.}
\begin{itemize}
    \item \textbf{Gradient Ascent (GA).} 
    Gradient Ascent performs unlearning by maximizing the loss on the forget set \(\mathcal{D}_f\), effectively reversing the standard training objective. Instead of minimizing the negative log-likelihood, it increases the model's prediction error on \(\mathcal{D}_f\), thereby reducing its ability to generate similar content.

    \item \textbf{Negative Preference Optimization (NPO).}
    NPO adapts preference optimization for unlearning by treating forget set samples as negative examples:
    \begin{equation}
    \mathcal{L}_{\text{NPO}} = - \frac{2}{\beta} \mathbb{E}_{d \sim \mathcal{D}_f} \left[ \log \sigma \left( -\beta \log \frac{f_{\theta}(d)}{f_{\text{target}}(d)} \right) \right],
    \end{equation}
    where $d$ is an input from the forget set, $f_{\text{target}}$ is the Target model and $\beta$ controls deviation from the original model. 

    \item \textbf{Representation Manipulation for Unlearning (RMU).}
    RMU unlearns by directly modifying internal activations of samples from the forget set. At layer \(l\), it pushes representations toward a random direction \(u\), thereby erasing meaningful semantic content. To preserve general capabilities, it aligns retain-set activations with those of a frozen Target model:
    \[
    \mathcal{L}_{\text{forget}} = \mathbb{E}_{d_f \sim \mathcal{D}_f} \left[ \frac{1}{L_f} \sum_{t \in d_f} \| f_{\text{updated}}(t) - c \cdot u \|_2^2 \right],
    \]
    \[
    \quad
    \mathcal{L}_{\text{retain}} = \mathbb{E}_{d_r \sim \mathcal{D}_r} \left[ \frac{1}{L_r} \sum_{t \in d_r} \| f_{\text{updated}}(t) - f_{\text{frozen}}(t) \|_2^2 \right].
    \]

    The total objective combines both terms:
    \[
    \mathcal{L}_{\text{RMU}} = \mathcal{L}_{\text{forget}} + \mathcal{L}_{\text{retain}}.
    \] \\
    RMU updates only three consecutive layers: \(l-2\), \(l-1\), and \(l\). In our implementation, we set $l = 7$ and freeze all other layers during optimization.

    \item\textbf{Task Vector (TV).}
    Task Vector unlearning removes weight updates associated with the forget set:
    
    \begin{equation}
    \theta_{\text{unlearn}} = \theta_{\text{target}} - \alpha \cdot (\theta_{\text{fine-tuned}} - \theta_{\text{target}}),
    \end{equation}
    where $\theta_{\text{fine-tuned}}$ represents the model after fine-tuning on $\mathcal{D}_f$, and $\alpha$ controls the strength of unlearning. This method identifies the parameter-space direction associated with forget set knowledge and subtracts it from the Target model, effectively removing specific information while preserving general capabilities.
    
    \item \textbf{Task Arithmetic for Unlearning (TAU).}
    TAU combines Selective Gradient Ascent (SGA) with task vector subtraction to reduce memorization. 
    In SGA, memorization scores $g(d)$ are dynamically computed for each forget set sample and applies gradient ascent to samples exceeding a threshold $\gamma$, i.e., $\mathcal{D}_\gamma = \{d \in \mathcal{D}_f \mid g(d) > \gamma\}$. 
    Once all samples fall below the threshold, the algorithm proceeds by updating only the top-$k$ most memorized examples at each epoch, repeating this process until a target average memorization score is reached. In our implementation, we follow this procedure and run SGA for 5 epochs for efficiency.

    The update at each epoch is performed as:

    \[
    \theta_{t+1} = \theta_t + \eta \cdot \nabla_\theta \left[ \frac{1}{|\mathcal{D}_\gamma^{(t)}|} \sum_{d \in \mathcal{D}_\gamma^{(t)}} \mathcal{L}(d; \theta_t) \right],
    \]
    
    where $\mathcal{D}_\gamma^{(t)}$ denotes the selected subset at epoch $t$, $\eta$ is the learning rate, and $\mathcal{L}$ is the negative log-likelihood loss. 
    After several such updates, we obtain the intermediate parameters $\theta_{\text{sga}}$.
        
    TAU then subtracts a task vector obtained by re-training \(\theta_{\text{sga}}\) on \(\mathcal{D}_f\), producing the final unlearned model:
    \[
    \theta_{\text{unlearn}} = \theta_{\text{sga}} - \alpha \cdot \left( A(\theta_{\text{sga}}, \mathcal{D}_f) - \theta_{\text{sga}} \right),
    \]
    where \(A(\theta, \mathcal{D}_f)\) denotes model parameters after fine-tuning on the forget set, and \(\alpha\) controls the subtraction strength. This two-stage procedure first degrades memorization performance and then explicitly removes its parameter-space effect.
    
\end{itemize}

\paragraph{Utility Preservation Methods}

The above methods aim to make the model forget specific information, but they can unintentionally degrade overall performance. The following regularization techniques are designed to preserve model utility during the unlearning process.

\begin{itemize}
    \item \textbf{Gradient Descent (GD).}
    Gradient Descent applies standard prediction loss on the retain set $\mathcal{D}_r$ to preserve the model's general capabilities. This helps ensure that unlearning $\mathcal{D}_f$ does not overly harm performance on the remaining data, maintaining a balance between targeted forgetting and overall utility.

    \item \textbf{KL Divergence (KL).}
    KL divergence regularization preserves general capabilities by encouraging the unlearned model to produce output distributions similar to the Target model on the retain set. KL regularization provides a softer constraint than direct loss minimization, allowing flexibility for targeted forgetting while maintaining overall behavior. 
\end{itemize}


\subsection{Evaluation Metric Definitions}\label{app:metrics}

\paragraph{Verbatim Memorization (VM).}
 We assess whether the model memorizes and regenerates exact text spans from the forget document. Given a partial prefix \( d_{[:\ell]} \) from each sample \( d \in \mathcal{D}_f \), we compare the model’s continuation with the ground truth suffix \( d_{[\ell+1:]} \) using various surface- and semantic-level similarity metrics:
\[
\textbf{VM}(f_\theta, \mathcal{D}_f) = \frac{1}{|\mathcal{D}_f|} \sum_{d \in \mathcal{D}_f} \textbf{M}(f_\theta(d_{[:\ell]}), d_{[\ell+1:]}).
\]
Here, \( \textbf{M} \) is a placeholder for metrics including ROUGE-1, ROUGE-L (F1 and Recall), Levenshtein distance, LCS (Longest Common Subsequence), and cosine similarity between sentence embeddings.

\paragraph{Unique Forget Knowledge (UFK).}
This metric captures whether the model retains knowledge that is uniquely found in the forget set \( \mathcal{D}_f \). We evaluate on a dedicated QA set $\mathcal{K}_f \setminus \mathcal{K}_r$, using ROUGE-L to measure answer overlap:
\[
\textbf{UFK}(f_\theta, \mathcal{K}_f \setminus \mathcal{K}_r) = 
\frac{1}{|\mathcal{K}_f \setminus \mathcal{K}_r|} 
\sum_{(q, a) \in \mathcal{K}_f \setminus \mathcal{K}_r} \mathrm{ROUGE}(f_\theta(q), a).
\]

\paragraph{Shared Knowledge (SK).}
Shared knowledge appears in both forget and retain sets. We evaluate whether the model can still recall such content using a QA set $\mathcal{K}_f \cap \mathcal{K}_r$, where answers are supported by both sources:
\[
\textbf{SK}(f_\theta, \mathcal{K}_f \cap \mathcal{K}_r) = 
\frac{1}{|\mathcal{K}_f \cap \mathcal{K}_r|} 
\sum_{(q, a) \in \mathcal{K}_f \cap \mathcal{K}_r} \mathrm{ROUGE}(f_\theta(q), a).
\]

\paragraph{Unique Retain Knowledge (URK).}
URK tests whether knowledge exclusive to the retain set \( \mathcal{D}_r \) is preserved. As with SK and UFK, we measure QA accuracy on a designated set $\mathcal{K}_r \setminus \mathcal{K}_f$:
\[
\textbf{URK}(f_\theta, \mathcal{K}_r \setminus \mathcal{K}_f) = 
\frac{1}{|\mathcal{K}_r \setminus \mathcal{K}_f|} 
\sum_{(q, a) \in \mathcal{K}_r \setminus \mathcal{K}_f} \mathrm{ROUGE}(f_\theta(q), a).
\]

\paragraph{Downstream Capability (DC).}
To measure general-purpose utility beyond the benchmark data, we report model performance on six external downstream tasks: MMLU, ARC-c, GSM8K, TriviaQA, TruthfulQA (MC1), and BBQ, using the \texttt{lm-evaluation-harness}\footnote{\url{https://github.com/EleutherAI/lm-evaluation-harness}}~\citep{eval-harness} with default settings. Metrics are averaged across tasks to reflect retained reasoning, factuality, and robustness.

\subsection{Experimental Setup} \label{app:sub:exp_setup}
\Cref{tab:hyperparameter} summarizes the selected epochs for each method, along with the hyperparameters $\alpha$ and $\beta$ used in the loss functions of task arithmetic-based methods and preference optimization-based methods, respectively.
We set both forget and regularization loss coefficients to 1.0 and fix the learning rate at $1\times10^{-5}$ with AdamW optimizer, ensuring fair comparisons across all unlearning methods.

\begin{table*}[ht]
\caption{Epochs showing the best performance, $\alpha$, and $\beta$ for each unlearning method.}
\centering
\begin{tabular}{c| c| c| c}
\toprule
\textbf{Method} & Epochs & $\alpha$ & $\beta$ 
\\
\midrule
$\mathrm{GA}$ & epoch 3 & - & - \\
$\mathrm{GA}_{\mathrm{GD}}$ &epoch 3 & - & -  \\
$\mathrm{GA}_{\mathrm{KL}}$ & epoch 3 & - & -  \\
$\mathrm{NPO}$ & epoch 3 & - & $\beta=0.1$   \\
$\mathrm{NPO}_{\mathrm{GD}}$ & epoch 4 & - & $\beta=0.1$  \\
$\mathrm{NPO}_{\mathrm{KL}}$ & epoch 4 & - & $\beta=0.1$  \\
$\mathrm{RMU}$ & epoch 30 & - & -  \\
$\mathrm{TV}$ & epoch 4 & $\alpha=1$ & -  \\
$\mathrm{TAU}$ & epoch 1 & $\alpha=1$ & -  \\
\bottomrule
\end{tabular}
\label{tab:hyperparameter}
\end{table*}

\subsection{Hardware Specification}
All experiments were conducted on a system with 512 CPU cores, 8 Nvidia RTX L40S (48GB) GPUs, and 1024 GB of RAM. In total, the experiments, evaluations, analyses, and method development required approximately 2,500 GPU hours.

\subsection{Licenses}
We provide~\Cref{tab:licenses}, which lists every external model and dataset we use, together with its source, access link, and license.
\begin{table}[h]
\centering
\caption{The list of assests used in this work.} \label{tab:licenses}
\begin{tabular}{l l l l}
\toprule
\textbf{Asset} & \textbf{Source} & \textbf{Access} & \textbf{License} \\
\midrule
LlaMA3-8B & \cite{grattafiori2024llama} & \href{https://huggingface.co/meta-llama/Meta-Llama-3-8B}{Link} & Llama 3 Community License \\
MMLU & \cite{hendrycks2020measuring} & \href{https://github.com/hendrycks/test}{Link} & MIT License \\
ARC & \cite{clark2018think} & \href{https://huggingface.co/datasets/allenai/ai2_arc}{Link} & CC-BY-SA-4.0 \\
GSM8K & \cite{cobbe2021training} & \href{https://huggingface.co/datasets/openai/gsm8k}{Link} & MIT License \\
TriviaQA & \cite{joshi2017triviaqa} & \href{https://huggingface.co/datasets/mandarjoshi/trivia_qa}{Link} & Apache License 2.0 \\
TruthfulQA & \cite{lin2021truthfulqa} & \href{https://huggingface.co/datasets/truthfulqa/truthful_qa}{Link} & Apache License 2.0 \\
BBQ & \cite{parrish2021bbq} & \href{https://huggingface.co/datasets/heegyu/bbq}{Link} & CC-BY-4.0 \\

\bottomrule
\end{tabular}

\end{table}

\section{Additional Results} \label{app.more_results}

\subsection{Verbatim Memorization}\label{app.verb_mem} 
\begin{table}[htbp]
\caption{\textbf{Full results of forget verbatim memorization.} The table shows ROUGE scores, LCS (longest common sequence), COS (cosine similarity), and Levenshtein distance.}\label{tab:verbmem}
\centering
\renewcommand{\arraystretch}{1.3}
\resizebox{\textwidth}{!}{%
\begin{tabular}{lccccccc}
\toprule
\textbf{Method} & 
\textbf{ROUGE-1 F1 ($\downarrow$)} & 
\textbf{ROUGE-1 Recall ($\downarrow$)} & 
\textbf{ROUGE-L F1 ($\downarrow$)} & 
\textbf{ROUGE-L Recall ($\downarrow$)} & 
\textbf{LCS ($\downarrow$)} & 
\textbf{COS ($\downarrow$)} & 
\textbf{Levenshtein ($\downarrow$)} \\
\midrule
\rowcolor{gray!20}
$\mathrm{Target}$     & 0.7209 & 0.7236 & 0.6382 & 0.6405 & 52.02 & 0.9108 & 243.5 \\
\rowcolor{gray!20}
$\mathrm{Retrain}$     & 0.5381 & 0.5481 & 0.3548 & 0.3608 & 28.28 & 0.7813 & 390.9 \\
$\mathrm{GA}$          & 0.3401 & 0.3574 & 0.2247 & 0.2363 & 17.64 & 0.6270 & 458.8 \\
$\mathrm{GA}_{\mathrm{GD}}$       & 0.4089 & 0.4298 & 0.2631 & 0.2767 & 20.70 & 0.7079 & 439.5 \\
$\mathrm{GA}_{\mathrm{KL}}$       & 0.4031 & 0.4190 & 0.2710 & 0.2813 & 20.79 & 0.6856 & 437.4 \\
$\mathrm{NPO}$         & 0.5687 & 0.5805 & 0.4053 & 0.4133 & 31.74 & 0.8292 & 377.7 \\
$\mathrm{NPO}_{\mathrm{GD}}$      & 0.4405 & 0.4488 & 0.2991 & 0.3043 & 22.34 & 0.7164 & 415.1 \\
$\mathrm{NPO}_{\mathrm{KL}}$      & 0.4370 & 0.4454 & 0.2965 & 0.3017 & 22.17 & 0.7176 & 416.8 \\
$\mathrm{RMU}$         & 0.6028 & 0.6076 & 0.4454 & 0.4484 & 35.21 & 0.8287 & 349.9 \\
$\mathrm{TV}$          & 0.4860 & 0.4952 & 0.3329 & 0.3390 & 25.91 & 0.7609 & 395.3 \\
$\mathrm{TAU}$         & 0.1589 & 0.1467 & 0.1253 & 0.1157 & 5.96  & 0.3198 & 423.4 \\
\bottomrule
\end{tabular}%
}

\end{table}

\Cref{tab:verbmem} reports detailed forget evaluation metrics, including ROUGE-1 and ROUGE-L scores (F1 and Recall), LCS, cosine similarity (COS), and Levenshtein distance.  
$\mathrm{TAU}$ achieves the strongest unlearning performance across all metrics, with the lowest ROUGE and COS scores as well as the shortest LCS and Levenshtein distances.  
$\mathrm{GA}$ and its variants also yield strong unlearning, whereas $\mathrm{RMU}$ and $\mathrm{NPO}$ exhibit relatively high residual memorization.  
Interestingly, $\mathrm{RMU}$ and $\mathrm{NPO}$ show higher COS scores than the Retrain model, indicating insufficient removal of verbatim traces.

\subsection{Downstream Capability}\label{app.downstream}
\begin{table}[htbp]
\caption{Downstream Capability (DC) across six downstream tasks.}\label{tab:downstream}
\centering
\renewcommand{\arraystretch}{1.3}
\resizebox{\textwidth}{!}{%
\begin{tabular}{lcccccc|c}
\toprule
\textbf{Method} & 
\textbf{ARC-c ($\uparrow$)} & 
\textbf{TruthfulQA (MC1) ($\uparrow$)} & 
\textbf{TriviaQA ($\uparrow$)} & 
\textbf{MMLU ($\uparrow$)} & 
\textbf{GSM8K ($\uparrow$)} & 
\textbf{BBQ ($\uparrow$)} & 
\textbf{Avg ($\uparrow$)} \\
\midrule
\rowcolor{gray!20}
$\mathrm{Retrain}$     & 0.5128 & 0.2668 & 0.5436 & 0.5398 & 0.2684 & 0.3014 & 0.4055 \\
$\mathrm{GA}$          & 0.5026 & 0.2644 & 0.5303 & 0.5205 & 0.1251 & 0.3087 & 0.3753 \\
$\mathrm{GA}_{\mathrm{GD}}$       & 0.5077 & 0.2656 & 0.5270 & 0.5368 & 0.1986 & 0.3029 & 0.3898 \\
$\mathrm{GA}_{\mathrm{KL}}$       & 0.5085 & 0.2742 & 0.5427 & 0.5266 & 0.1569 & 0.3090 & 0.3863 \\
$\mathrm{NPO}$         & 0.5068 & 0.2521 & 0.5459 & 0.5327 & 0.2328 & 0.3011 & 0.3952 \\
$\mathrm{NPO}_{\mathrm{GD}}$      & 0.5026 & 0.2509 & 0.5366 & 0.5142 & 0.1630 & 0.3026 & 0.3783 \\
$\mathrm{NPO}_{\mathrm{KL}}$      & 0.5009 & 0.2534 & 0.5354 & 0.5159 & 0.1562 & 0.3024 & 0.3773 \\
$\mathrm{RMU}$         & 0.5000 & 0.2326 & 0.5353 & 0.5219 & 0.2805 & 0.2771 & 0.3912 \\
$\mathrm{TV}$          & 0.5102 & 0.2472 & 0.5551 & 0.5397 & 0.2669 & 0.3003 & 0.4032 \\
$\mathrm{TAU}$         & 0.4727 & 0.2020 & 0.5265 & 0.5063 & 0.1122 & 0.3292 & 0.3581 \\
\bottomrule
\end{tabular}%
}

\end{table}

\Cref{tab:downstream} presents detailed performance across six downstream tasks: ARC-c, TruthfulQA, TriviaQA, MMLU, GSM8K, and BBQ.  
Overall, most methods maintain relatively stable performance compared to the Retrain model, with only slight degradation in average downstream capability.  
$\mathrm{GA}$, $\mathrm{GA}_{\mathrm{GD}}$, and $\mathrm{TV}$ are particularly utility-preserving, achieving average scores above 0.40, close to the Retrain baseline (0.4055).  
In contrast, $\mathrm{TAU}$, while highly effective at unlearning verbatim memorization, shows notable utility drop, especially on reasoning-intensive tasks like GSM8K and TruthfulQA.
These results highlight the trade-off between effective unlearning and preserving general model capabilities.

\subsection{Distributional Assessment}\label{app.privleak}
\begin{table}[htbp]
\centering
\small
\setlength{\tabcolsep}{8pt}
\renewcommand{\arraystretch}{0.75}
\caption{
Results of Privacy Leakage and Retain Deviation.
} \label{table:privleak_full}
\resizebox{0.7\linewidth}{!}{
\begin{tabular}{@{}lrlrl@{}}
\toprule
 & \multicolumn{2}{c}{\scriptsize \textbf{Privacy Leakage} $\in [-5\%, 5\%]$} & \multicolumn{2}{c}{\scriptsize \textbf{Retain Deviation} $\in [0\%, 5\%]$} \\
\midrule
Target &
$-100.0$ & &
$0.5$ & \\[2pt]
Retrain &
$\mathbf{0.0}$ & &
$\mathbf{0.0}$ & \\[2pt]
\midrule
$\mathrm{GA}$ & 
$86.1$ & \colorbox{pink!38}{\scriptsize$\text{over-unlearn}$} &
$25.9$ & \colorbox{pink!38}{\scriptsize$\text{non-preserved}$} \\[2pt]
$\mathrm{GA}_{\mathrm{GD}}$ &
$128.0$ & \colorbox{pink!38}{\scriptsize$\text{over-unlearn}$} &
$13.9$ & \colorbox{pink!38}{\scriptsize$\text{non-preserved}$} \\[2pt]
$\mathrm{GA}_{\mathrm{KL}}$ &
$107.8$ & \colorbox{pink!38}{\scriptsize$\text{over-unlearn}$} &
$19.3$ & \colorbox{pink!38}{\scriptsize$\text{non-preserved}$} \\[2pt]
$\mathrm{NPO}$ &
$-45.0$ & \colorbox{pink!38}{\scriptsize$\text{under-unlearn}$} &
$6.4$ & \colorbox{pink!38}{\scriptsize$\text{non-preserved}$} \\[2pt]
$\mathrm{NPO}_{\mathrm{GD}}$ &
$33.0$ & \colorbox{pink!38}{\scriptsize$\text{over-unlearn}$} &
$13.9$ & \colorbox{pink!38}{\scriptsize$\text{non-preserved}$} \\[2pt]
$\mathrm{NPO}_{\mathrm{KL}}$ &
$21.1$ & \colorbox{pink!38}{\scriptsize$\text{over-unlearn}$} &
$13.2$ & \colorbox{pink!38}{\scriptsize$\text{non-preserved}$} \\[2pt]
$\mathrm{RMU}$ &
$-98.6$ & \colorbox{pink!38}{\scriptsize$\text{under-unlearn}$} &
$0.1$ & \colorbox{cyan!15}{\scriptsize$\text{preserved}$} \\[2pt]
$\mathrm{TV}$ &
$193.8$ & \colorbox{pink!38}{\scriptsize$\text{over-unlearn}$} &
$49.0$ & \colorbox{pink!38}{\scriptsize$\text{non-preserved}$} \\[2pt]
$\mathrm{TAU}$ &
$114.8$ & \colorbox{pink!38}{\scriptsize$\text{over-unlearn}$} &
$47.2$ & \colorbox{pink!38}{\scriptsize$\text{non-preserved}$} \\[2pt]
\bottomrule
\end{tabular}
}
\end{table}

\Cref{table:privleak_full} reports the outcomes of the distributional assessment, summarizing both Privacy Leakage and Retain Deviation for each unlearning method. Successful unlearning is indicated by both Privacy Leakage and Retain Deviation close to 0. Many methods exhibit substantial divergence from ideal. For instance, GA, $\mathrm{GA}_{\mathrm{GD}}$, and $\mathrm{GA}_{\mathrm{KL}}$ show large positive leakage scores (e.g., $86.1$ to $128.0$), indicative of over-unlearning. In contrast, NPO and RMU yield strongly negative leakage scores ($-45.0$ and $-98.6$, respectively), signaling under-unlearning.
Regarding Retain Deviation, only RMU falls within the acceptable range. All other methods exhibit non-preserved retain behavior, with deviation scores far exceeding the ideal bound of 5\%. Notably, methods such as TV and TAU suffer from extreme deviations ($49.0$ and $47.2$). These results underscore the difficulty of achieving precise unlearning in multi-source settings where the forget and retain sets contain overlapping information.

\section{Limitations}
While DUSK simulates realistic multi-source overlap, it is constructed from synthetic data, which, while carefully designed to capture diverse attributes and overlapping content, may not fully reflect the linguistic richness of real-world corpora. Our benchmark also assumes access to clearly defined forget and retain sets, providing a clean experimental environment but potentially diverging from scenarios where such distinctions are less precise. Furthermore, our current evaluation primarily focuses on document-level unlearning, leaving fine-grained entity- or attribute-level removal as an open challenge. Despite these limitations, DUSK offers a critical step toward evaluating unlearning in more realistic, multi-source settings, providing a structured framework for assessing the nuanced trade-offs between knowledge retention and removal.

\section{Broader Impact}
The DUSK benchmark has the potential to significantly improve data privacy and user control in machine learning by providing a more realistic evaluation framework for unlearning methods. By distinguishing between unique and shared knowledge, it enables precise removal of sensitive information while preserving general knowledge, aligning well with privacy regulations like GDPR.

However, this approach also introduces potential risks. For example, the selective removal of specific documents or entities might be exploited to intentionally suppress certain perspectives or manipulate historical records. Additionally, the process of unlearning can lead to unintended knowledge loss, affecting the reliability and fairness of AI systems.

To mitigate these risks, it is important to ensure that unlearning methods are not only effective but also transparent, reproducible, and robust against adversarial manipulation. Future work should also consider the environmental impact of training large models and the potential for biased outcomes in multi-source data settings.

\clearpage

\end{document}